\documentclass[letterpaper]{article} 
\usepackage{aaai2027}  
\usepackage[hyphens]{url}  
\usepackage{graphicx} 
\usepackage{natbib}  
\usepackage{caption} 
\usepackage{amsmath}
\usepackage{amssymb}
\usepackage{algorithm}
\usepackage{algorithmic}

\usepackage[table]{xcolor}
\definecolor{groupbg}{RGB}{232,235,238}

\usepackage{newfloat}
\usepackage{listings}
\DeclareCaptionStyle{ruled}{labelfont=normalfont,labelsep=colon,strut=off} 
\floatstyle{ruled}
\newfloat{listing}{tb}{lst}{}
\floatname{listing}{Listing}

\usepackage{booktabs}

\title{OPLD: On-Policy Latent Distillation for Multimodal Reasoning}
\author{
    Shoutai Zhu\textsuperscript{\rm 1},
    Tianyang Xu\textsuperscript{\rm 1},
    Sun Bin\textsuperscript{\rm 1},
    Xumingyuan\textsuperscript{\rm 1},
    Yu Liu\textsuperscript{\rm 1},
    Qinzhen Guo\textsuperscript{\rm 1}
}
\affiliations{
    \textsuperscript{\rm 1}ByteDance\\
    shoutaizhu61@gmail.com\\
    \{xutianyang.666, sunbin.824, xumingyuan.0916, liuyu.96, guoqinzhen\}@bytedance.com
}

\begin{document}

\maketitle

\begin{abstract}
Interleaved multimodal Chain-of-Thought (CoT) improves visual reasoning by incorporating auxiliary visual evidence into intermediate reasoning. However, existing approaches remain constrained by externally defined reasoning traces and visual operations, limiting their ability to develop flexible and abstract visual thinking. Reasoning with latent has recently offered a promising direction by internalizing intermediate computation into continuous representations. Nevertheless, existing visual-latent methods mainly supervise latent states through alignment with compressed auxiliary visual features, treating them as proxies for visual observations rather than active reasoning states. Consequently, they capture the provided evidence but fail to fully internalize the abstract reasoning process induced by multimodal CoT.

In this paper, we propose \textbf{OPLD} (\textbf{O}n-\textbf{P}olicy \textbf{L}atent \textbf{D}istillation), a simple framework that transfers the reasoning capability induced by privileged multimodal CoT into latent reasoning representations. During training, a CoT-augmented teacher observes privileged multimodal reasoning signals, while a CoT-free student receives the original input. The student first performs latent reasoning under its current policy, after which the teacher supervises the same reasoning trajectory through token-level distribution alignment and latent trajectory distillation. Consequently, the student internalizes multimodal reasoning patterns into latent space instead of merely imitating auxiliary visual features, enabling efficient reasoning without explicit CoT or auxiliary images during inference.
Extensive experiments on diverse multimodal benchmarks demonstrate that OPLD consistently outperforms existing latent reasoning methods and achieves state-of-the-art performance on multiple benchmarks. The results suggest that supervising latent representations at the reasoning-process level provides a more effective paradigm for multimodal latent reasoning than conventional feature-level alignment.
\end{abstract}


\begin{figure}[t]
\centering
\includegraphics[width=1.0\columnwidth]{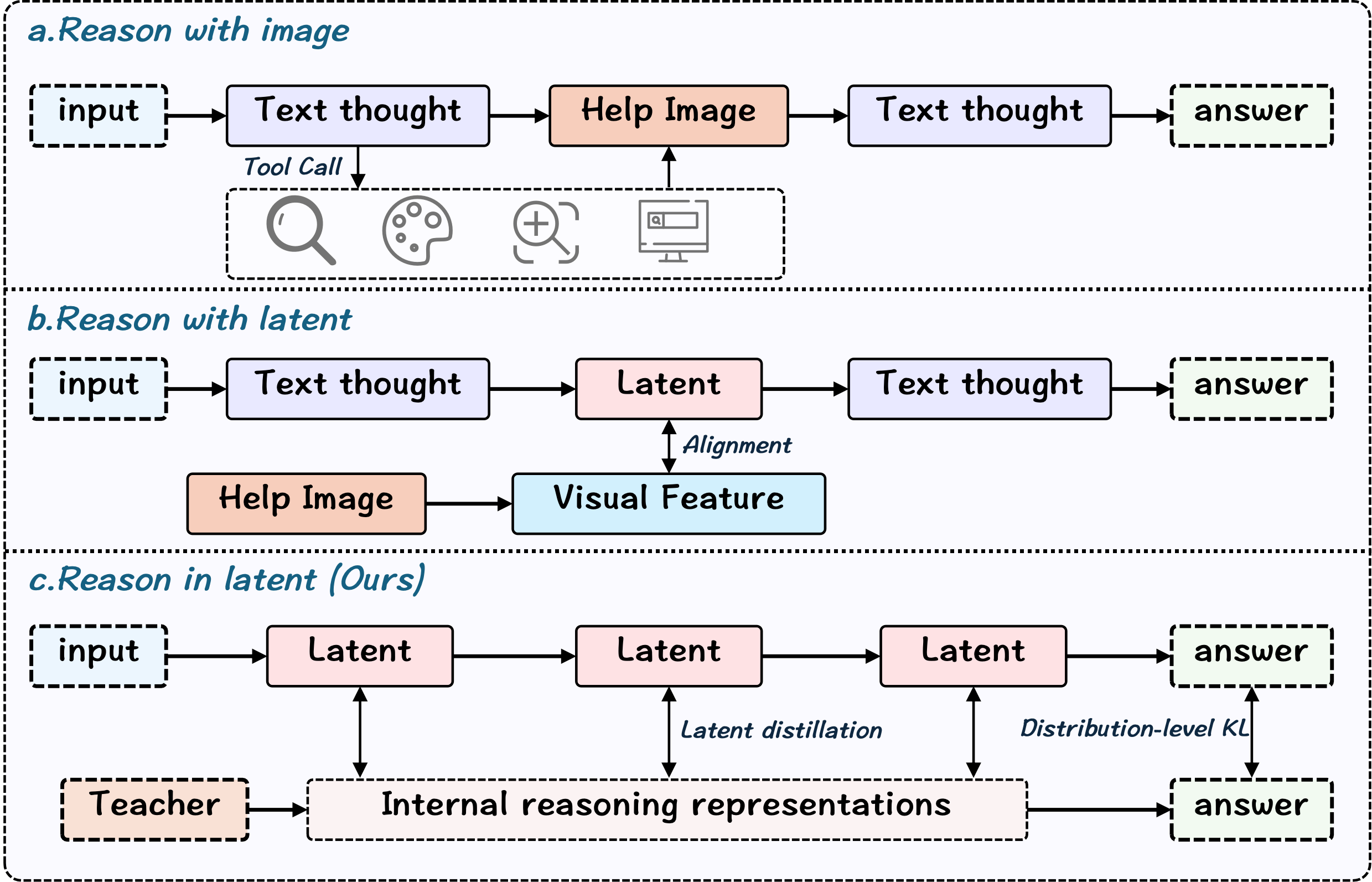}
\caption{
Comparison of multimodal reasoning paradigms.
OPLD learns latent reasoning by distilling privileged multimodal reasoning representations, eliminating the need for auxiliary images or textual CoT during inference.
}
\label{fig1}
\end{figure}

\section{Introduction}

Chain-of-thought reasoning has become an effective mechanism for improving the reasoning ability of large language models and multimodal large language models ~\cite{wei2022chain,kojima2022large,zhang2023multimodal}. For visual reasoning, recent work further shows that CoT need not be restricted to text: injecting visual evidence into intermediate reasoning steps can substantially improve the ability of MLLMs to solve perception-intensive and spatially grounded problems. This gives rise to interleaved multimodal CoT, where reasoning is supported not only by textual rationales but also by auxiliary images, such as cropped regions, zoomed-in visual evidence, annotated diagrams, or tool-generated observations. Existing methods mainly follow two technical paradigms. The first explicitly constructs visual reasoning traces through multi-step interaction, region prediction, visual tool invocation, or image manipulation, allowing the model to ``think with images'' during inference~\cite{zheng2025deepeyes,zhang2025thinkingimages,xia2025gcot}. The second performs reasoning in latent visual space: instead of explicitly materializing all auxiliary visual observations, some methods train MLLMs to generate continuous latent states that represent key visual evidence or intermediate visual thoughts~\cite{li2025latent,wang2025monet}. Both paradigms demonstrate the importance of visual intermediate reasoning, but they also reveal a key limitation: the reasoning process is either externalized as explicit CoT, tool-generated observations or specialized toward representing auxiliary visual features.

In this paper, we take a different perspective.
We ask whether latent states can learn the abstract internal reasoning representations that an MLLM forms when it leverages interleaved multimodal CoT. 
Latent reasoning provides a natural substrate for this goal. Prior work has shown that language and vision-language models can perform intermediate computation through continuous hidden states, from hidden-state feedback in LLMs~\cite{hao2024coconut} to latent visual reasoning in MLLMs~\cite{wang2025monet,jeon2026valr,li2025latent}. 
However, a central challenge remains: latent reasoning states lack explicit supervision. Existing methods often construct surrogate targets by aligning latent tokens with compressed auxiliary visual features. While convenient, this introduces several limitations, including information loss, representation mismatch, and train--inference inconsistency. Moreover, strong feature-level alignment can over-constrain the model's internal states, making it prone to catastrophic forgetting, shortcut learning, and task-specific overfitting. These issues limit the generality of latent reasoning methods. In contrast, we seek to supervise latent states at the reasoning-process level, encouraging them to capture the abstract internal representations induced by interleaved multimodal CoT rather than merely imitating auxiliary visual features.

We propose \textbf{OPLD}, an \textbf{On-Policy Latent Distillation} framework for multimodal latent reasoning. OPLD consists of a teacher model and a student model with same MLLM backbone and latent reasoning structure, but different input conditions. The teacher is equipped with interleaved multimodal CoT, enabling it to form stronger latent reasoning trajectories under guidance. The student is equipped with only original image-question input, encouraging it to develop a CoT-free latent reasoning process without relying on explicit rationale tokens. Training proceeds in three stages. We first train the CoT-guided teacher, then warm up the CoT-free student to establish a stable latent reasoning protocol, and finally perform on-policy latent distillation. In the distillation stage, the student generates answers from its current policy, and the teacher evaluates these student-generated trajectories to provide token-level supervision. Meanwhile, the student's latent trajectory is aligned with the teacher's latent trajectory, allowing the abstract reasoning patterns induced by multimodal CoT to be transferred into the student's latent space. 
Consequently, the final student model internalizes the abstract reasoning patterns of multimodal CoT into latent space, enabling it to answer through latent reasoning and imagination in the absence of auxiliary information.

Our contributions are summarized as follows:
\begin{itemize}
    \item We propose \textbf{OPLD}, a simple and general on-policy latent distillation framework for multimodal latent reasoning. OPLD does not require task-specific visual tools or explicit auxiliary reasoning at inference time. Instead, it internalizes the abstract reasoning patterns induced by textual and interleaved multimodal CoT into latent space.

    \item We introduce an on-policy teacher--student paradigm for latent reasoning. The student rolls out answers with its current policy, and the CoT-augmented teacher supervises the same trajectories through token-level feedback and latent trajectory alignment, enabling the student to answer via latent reasoning without auxiliary information.

    \item OPLD improves both multimodal perception and reasoning. Extensive experiments show that OPLD consistently outperforms strong latent reasoning methods and achieves state-of-the-art performance on multiple benchmarks.
\end{itemize}

\section{Related Work}

\subsection{Reasoning with Images}

CoT prompting improves complex reasoning by decomposing the answer into intermediate steps~\cite{wei2022chain,kojima2022large}. Early multimodal CoT methods extend this idea to vision-language tasks by generating textual rationales conditioned on both the image and the question~\cite{zhang2023multimodal,chen2024vctp,shao2024viscot}. However, text-only rationales can be insufficient for visual reasoning, since the model must often revisit fine-grained regions, verify spatial relations, read small text, or compare local visual details. This motivates a growing body of work on visual or grounded CoT, where intermediate reasoning steps are explicitly tied to visual evidence. Representative methods guide MLLMs to extract visual rationales step by step, ground reasoning steps to image regions, detect relevant regions before answering, or replay cropped visual evidence during generation~\cite{zhou2024iot,wu2025gcot,man2025argus,wang2025vgr,jiang2025vlmr3,hu2026tvicot}. These methods show that exposing task-relevant visual evidence during intermediate reasoning can substantially improve perception-intensive and spatially grounded tasks.

Beyond grounded rationales, recent work further explores the thinking with images paradigm, where vision becomes an active workspace rather than a static input. Some methods perform multi-round visual exploration by predicting where to crop, zoom, or revisit, sometimes optimized with reinforcement learning or region-level rewards~\cite{jiang2025vlmr3,zheng2025deepeyes,liu2026efficienticot}. Others introduce generative visual reasoning, where models produce intermediate visual thoughts, subgoal images, RGB visual intermediates, or edited visual states to support downstream reasoning and planning~\cite{li2025mvot,chern2025thinkinggenerated,zhou2026genvcot,zhao2025cotvla,yin2026gvcot}. These approaches provide a powerful and interpretable way to incorporate visual evidence into the reasoning process, but they also share a common limitation: the intermediate reasoning process remains externalized. Inference may require explicit rationales, region selection, tool invocation, image generation, or additional visual observations, which introduces error propagation and operational rigidity. In contrast, OPLD uses interleaved multimodal CoT as training guidance and internalizes its abstract reasoning patterns into latent computation, enabling the model to reason without auxiliary visual traces at inference time.

\subsection{Reasoning with Latents}

Latent reasoning aims to move intermediate computation from explicit token sequences into continuous hidden space. In language models, recent work shows that hidden states can be recurrently fed back as continuous thoughts, allowing models to reason beyond discrete chain-of-thought tokens~\cite{hao2024coconut}. This idea has been extended to multimodal reasoning, where continuous latents are used to represent visual thoughts, mental imagery, or intermediate visual states. LVR performs autoregressive reasoning directly in the visual embedding space by training latent states to reconstruct key visual tokens relevant to the query~\cite{li2025latent}. Mirage augments VLM decoding with latent visual tokens interleaved with text tokens, using image-embedding distillation and subsequent task supervision to support multimodal mental imagery~\cite{yang2025mirage}. Monet further enables MLLMs to generate continuous embeddings as intermediate visual thoughts and introduces visual-latent policy optimization to explicitly optimize latent reasoning~\cite{wang2025monet}. SkiLa proposes latent sketch tokens that alternate with textual thinking tokens, enabling unified text-visual reasoning in a shared latent space~\cite{tong2025skila}. Laser improves visual deduction through dynamic windowed alignment, allowing latent states to maintain coarse-to-fine semantic superposition instead of enforcing rigid point-wise prediction~\cite{wang2026laser}. HyLaR formulates multimodal reasoning as a hybrid discrete-continuous action space and applies decoupled policy optimization to jointly train textual and latent actions~\cite{cheng2026hylar}.

However, supervising latent reasoning remains intrinsically difficult because latent states do not have natural ground-truth annotations. Existing methods therefore often construct surrogate targets by aligning latents with compressed image embeddings, key visual tokens, latent sketches, or intermediate visual observations~\cite{li2025latent,yang2025mirage,wang2025monet,tong2025skila}. Although effective, such supervision largely treats latent states as proxies for auxiliary visual features. This can introduce information loss, representation mismatch, and train--inference inconsistency, while strong feature-level alignment may over-constrain internal representations and reduce generality. Recent diagnostic studies further show that latent tokens may be weakly coupled with both the visual input and the final answer, or become semantically enriched yet under-utilized during answer prediction~\cite{li2026imagination,zhang2026unsilencing}. These findings suggest that the key issue is not merely how to generate latent visual tokens, but how to make them function as active reasoning states.

OPLD takes a different perspective. Rather than forcing latent states to imitate auxiliary visual features, it learns latent trajectories from the abstract reasoning patterns induced by textual and interleaved multimodal CoT. In this way, OPLD encourages latent states to support answer generation as internal reasoning abstractions, enabling the final model to answer through latent reasoning and imagination without auxiliary information at inference time.

\begin{figure*}[t]
\centering
\includegraphics[width=0.9\textwidth]{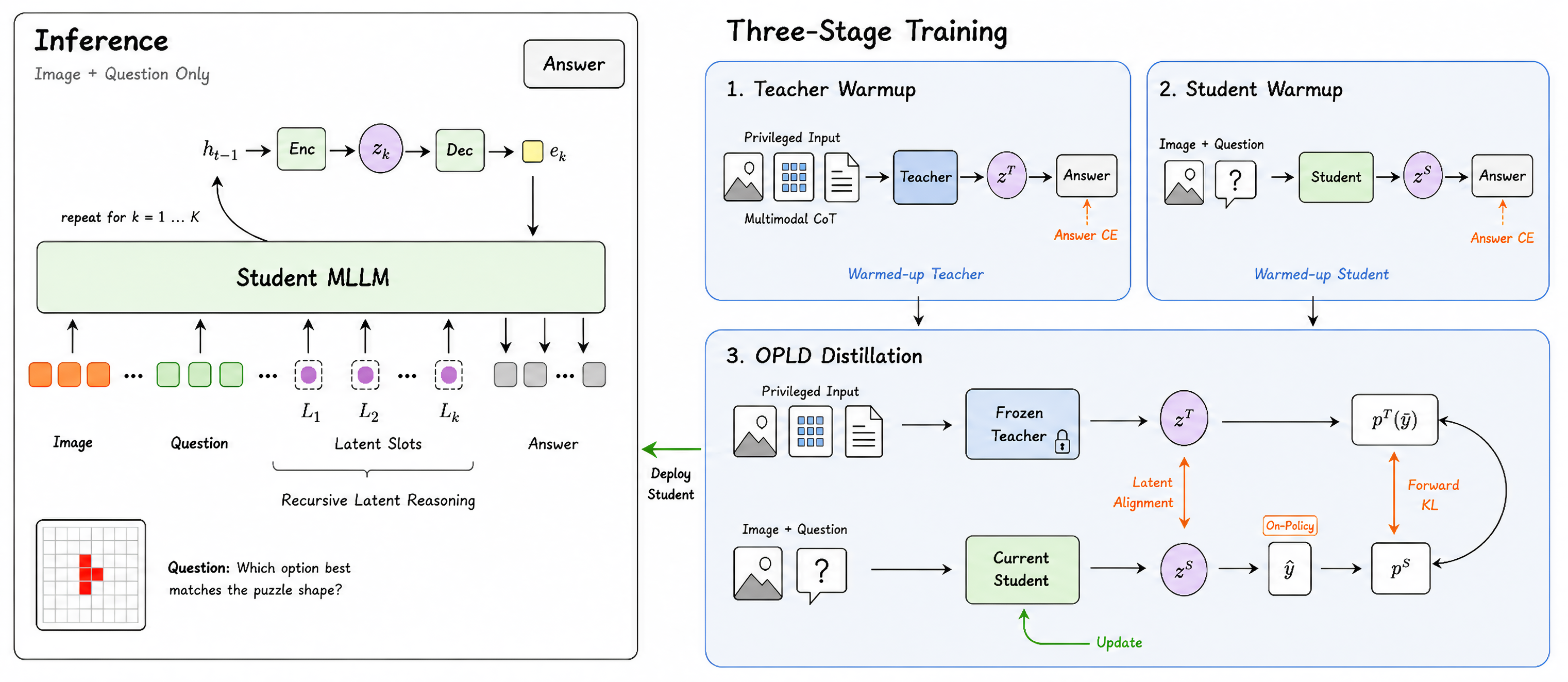}
\caption{
Overview of OPLD. 
The left shows inference, where the model answers from the original image and question through recursive latent reasoning. 
The right illustrates the three-stage training pipeline: a CoT-guided teacher is first warmed up with interleaved multimodal CoT, a CoT-free student is then warmed up with only the image-question input, and finally the student is optimized by on-policy latent distillation using token-level feedback and latent trajectory alignment from the frozen teacher.
}
\label{method}
\end{figure*}

\section{Method}

We propose \textbf{OPLD}, an on-policy latent distillation framework for multimodal latent reasoning. As shown in Fig.~\ref{method}, OPLD contains a teacher model and a student model with the same MLLM backbone and recursive latent reasoning structure, but different input conditions. The teacher reasons with interleaved multimodal CoT, including textual rationales and auxiliary visual evidence, while the student receives only the original image-question input. The objective of OPLD is to transfer the abstract reasoning patterns induced by multimodal CoT into the student's latent space, so that the final model can answer through latent reasoning and imagination without auxiliary information at inference time.

OPLD follows a three-stage training pipeline. First, we train a CoT-guided teacher with auxiliary multimodal CoT, enabling it to form strong latent reasoning trajectories. Second, we warm up a CoT-free student using only the original image and question, establishing a stable latent reasoning protocol before distillation. Third, we perform on-policy latent distillation. In this stage, the student first generates answers with its current policy, and the frozen teacher supervises the same student-generated trajectories through token-level feedback and latent trajectory alignment. At inference time, the teacher, auxiliary CoT, help images, and all other auxiliary information are discarded. The student performs a fixed number of recursive latent reasoning steps and then generates the answer from the original image and question alone.

\subsection{Problem Setup}

For each training example, we define two input views. The student view contains only the original image and question:
\begin{equation}
x_S = (I, Q),
\end{equation}
where \(I\) denotes the main image and \(Q\) denotes the question. The teacher view additionally contains interleaved multimodal CoT:
\begin{equation}
x_T = (I, Q, C), \quad C = (R, I_{\mathrm{help}}),
\end{equation}
where \(R\) denotes textual CoT or textual evidence, and \(I_{\mathrm{help}}\) denotes auxiliary help images, such as cropped regions, zoomed-in visual evidence, annotated visual rationales, or other task-relevant auxiliary images. The teacher view is used only during training, while the deployed model always receives \(x_S\).

Let
\begin{equation}
y^{\ast} = (y^{\ast}_1, \ldots, y^{\ast}_N)
\end{equation}
denote the ground-truth answer. In the first two warmup stages, \(y^{\ast}\) is used for answer-only supervised training. In the third distillation stage, the current student generates an on-policy response:
\begin{equation}
\hat{y} = (\hat{y}_1, \ldots, \hat{y}_M),
\end{equation}
which is then evaluated by the teacher. Moreover, we denote the recursive latent trajectories of the student and teacher as
\begin{equation}
z^S_{1:K} = (z^1_S, \ldots, z^K_S),
\quad
z^T_{1:K} = (z^1_T, \ldots, z^K_T),
\end{equation}
where \(K\) is the number of latent reasoning slots. Both the teacher and the student use the native MLLM chat template. We append \(K\) latent slots between the assistant generation prompt and answer generation.

\subsection{Recursive Latent Reasoning}

OPLD performs intermediate reasoning through a fixed number of recursive latent slots. 
Rather than directly reusing raw LLM hidden states as latent thoughts, we project them into a lower-dimensional latent space. 
This design creates a compact abstraction and encourages the latent states to capture the abstract internal reasoning representations induced by multimodal CoT. 
A lightweight decoder then maps each latent code back to the MLLM embedding space, so the model can consume the latent state without changing the backbone architecture.

Let \(p_k\) be the position of the \(k\)-th latent slot, and let \(h^L_{p_k-1}\in\mathbb{R}^{H}\) be the last-layer hidden state at the position immediately before this latent slot. OPLD computes the latent code and the corresponding input embedding as
\begin{equation}
z_k = \mathrm{Enc}(h^L_{p_k-1}),
\quad
e_k = \mathrm{Dec}(z_k).
\end{equation}
Here, \(z_k\in\mathbb{R}^{D}\) is the latent reasoning code in the projected latent space, and \(e_k\in\mathbb{R}^{H}\) is the decoded embedding written into the \(k\)-th latent slot. The encoder and decoder are lightweight MLP projection modules:
\begin{equation}
\mathrm{Enc}(h) =
\mathrm{LN}
\left(
W_2 \, \mathrm{GELU}(W_1 \, \mathrm{LN}(h))
\right),
\end{equation}
\begin{equation}
\mathrm{Dec}(z) =
W_4 \, \mathrm{GELU}(W_3 \, \mathrm{LN}(z)).
\end{equation}

The latent trajectory is produced recursively through hidden-state feedback.
For the first latent slot, the previous position corresponds to the assistant generation prompt:
\begin{equation}
z_1 = \mathrm{Enc}(h^L_{p_1-1}),
\quad
e_1 = \mathrm{Dec}(z_1).
\end{equation}
For subsequent latent slots, the model re-runs the forward computation with the decoded embeddings from earlier latent steps inserted into their corresponding positions. The next latent code is then computed as
\begin{equation}
z_k = \mathrm{Enc}(h^L_{p_k-1}),
\quad
e_k = \mathrm{Dec}(z_k),
\qquad k=2,\ldots,K.
\end{equation}
After computing \(z_k\), the decoded embedding \(e_k = \mathrm{Dec}(z_k)\) is written into the last latent slot. Therefore, later latent states depend on earlier latent states, forming a fixed-step internal reasoning chain. This recursive process enables the model to accumulate abstract reasoning states before producing any answer token. The mechanism is shared by the teacher and the student. During inference, the student first executes this \(K\)-step latent reasoning process and then generates answer tokens.

\subsection{Three-Stage OPLD Training}

OPLD contains three stages: teacher warmup, student warmup, and on-policy latent distillation. The first two stages establish stable latent reasoning dynamics for the teacher and the student, while the third stage transfers CoT-induced reasoning from the teacher into the student's latent space.

\paragraph{Stage 1: Teacher Warmup.}
The first stage trains a CoT-guided teacher using the teacher view \(x_T = (I,Q,C)\). Since the teacher has access to textual CoT and auxiliary help images, it can learn latent trajectories guided by richer multimodal reasoning evidence. Given the teacher latent trajectory \(z^T_{1:K}\), we optimize the teacher with answer-only cross-entropy:
\begin{equation}
\mathcal{L}_{\mathrm{T}}
=
-\sum_{n=1}^{N}
\log p_{\phi}
\left(
y^{\ast}_n
\mid
x_T,z^T_{1:K},y^{\ast}_{<n}
\right),
\end{equation}
where \(\phi\) denotes the teacher parameters. Only answer tokens are supervised, while prompt tokens and latent slots are masked. This stage is not used for distillation but to form a teacher whose latent trajectory reflects the abstract reasoning patterns induced by interleaved multimodal CoT.

\paragraph{Stage 2: Student Warmup.}
The second stage trains a CoT-free student using only the student view \(x_S = (I,Q)\). The student does not observe textual CoT or help images. Given the student latent trajectory \(z^S_{1:K}\), we apply the same answer-only cross-entropy:
\begin{equation}
\mathcal{L}_{\mathrm{S}}
=
-\sum_{n=1}^{N}
\log p_{\theta}
\left(
y^{\ast}_n
\mid
x_S,z^S_{1:K},y^{\ast}_{<n}
\right),
\end{equation}
where \(\theta\) denotes the student parameters. This warmup stage allows the student to establish a stable CoT-free latent reasoning protocol before distillation. Without this stage, the student would enter distillation with poorly formed latent dynamics, making latent trajectory alignment unstable.

\begin{figure*}[t]
\centering
\includegraphics[width=0.9\textwidth]{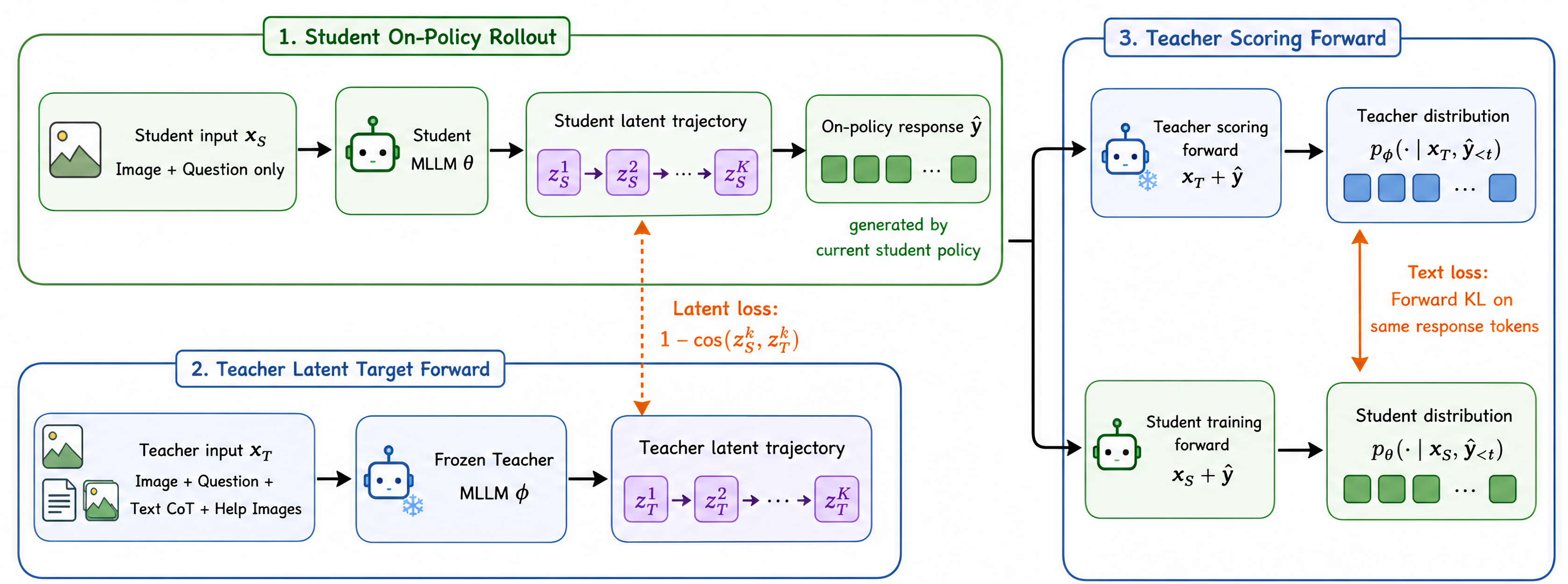}
\caption{
OPLD distillation stage.
The student rolls out response \(\hat{y}\) with its current policy from the original image-question input.
The frozen CoT-guided teacher provides two forms of supervision on this student-generated trajectory: latent alignment between teacher and student latent trajectories, and token-level forward KL on the same response tokens.
}
\label{opld}
\end{figure*}

\paragraph{Stage 3: OPLD Distillation.}
After warmup, the teacher is frozen and only the student is updated. As detailed in Fig.~\ref{opld}, OPLD distillation consists of three forward processes: student on-policy rollout, teacher latent target forward, and teacher scoring forward.

First, the current student performs an on-policy rollout. It receives only \(x_S\), produces its recursive latent trajectory, and generates an answer:
\begin{equation}
z^S_{1:K} = f^{\mathrm{lat}}_{\theta}(x_S),
\quad
\hat{y} \sim \pi_{\theta}(\cdot \mid x_S,z^S_{1:K}).
\end{equation}

Second, the frozen teacher performs a latent target forward under the teacher view:
\begin{equation}
z^T_{1:K} = f^{\mathrm{lat}}_{\phi}(x_T).
\end{equation}
The teacher latent trajectory provides the CoT-guided target for the student's latent reasoning. We align the student and teacher latent trajectories using cosine distance:
\begin{equation}
\mathcal{L}_{\mathrm{lat}}
=
\frac{1}{BK}
\sum_{b=1}^{B}
\sum_{k=1}^{K}
\left[
1 -
\cos
\left(
z^k_{S,b},
\mathrm{sg}(z^k_{T,b})
\right)
\right],
\end{equation}
where \(B\) is the batch size and \(\mathrm{sg}(\cdot)\) denotes stop-gradient. The teacher latent states are treated as fixed targets, and gradients are applied only to the student. This loss encourages the student latent trajectory to approximate the abstract reasoning trajectory formed under multimodal CoT guidance.


Third, the teacher and student score the same student-generated trajectory. 
Different from the latent target forward, the teacher scoring forward does not use the its own latent trajectory. 
Instead, it takes the student rollout, including both the student latent codes \(z^S_{1:K}\) and the student-generated response \(\hat{y}\). 
This design lets the CoT-guided teacher provide token-level feedback on the student's on-policy trajectory.

For each response position \(t\), the teacher distribution is computed under the privileged teacher view while conditioning on the student latent rollout:
\begin{equation}
p^T_t
=
p_{\phi}
\left(
\cdot
\mid
x_T,z^S_{1:K},\hat{y}_{<t}
\right),
\end{equation}
and the student distribution is computed under the student view:
\begin{equation}
p^S_t
=
p_{\theta}
\left(
\cdot
\mid
x_S,z^S_{1:K},\hat{y}_{<t}
\right).
\end{equation}
Therefore, both models are evaluated on the same trajectory while differing only in their information views. 
The resulting distribution gap provides a direct token-level supervision signal for improving the student's behavior. We use top-\(k\) forward KL as the text-level distillation loss.
For each position \(t\), let
\begin{equation}
\mathcal{V}^T_t = \mathrm{TopK}(p^T_t)
\end{equation}
denote the set of tokens with the highest teacher probabilities. 
The text-level distillation loss is defined as
\begin{equation}
\mathcal{L}_{\mathrm{text}}
=
\frac{1}{|\hat{y}|}
\sum_{t=1}^{|\hat{y}|}
\sum_{v\in\mathcal{V}^T_t}
p^T_t(v)
\left[
\log p^T_t(v)
-
\log p^S_t(v)
\right].
\end{equation}

The final OPLD objective combines token-level on-policy distillation and latent trajectory alignment:
\begin{equation}
\mathcal{L}_{\mathrm{OPLD}}
=
\lambda_{\mathrm{text}}\mathcal{L}_{\mathrm{text}}
+
\lambda_{\mathrm{lat}}\mathcal{L}_{\mathrm{lat}}.
\end{equation}
Through \(\mathcal{L}_{\mathrm{text}}\), the student learns from teacher feedback on its own generated answers. Through \(\mathcal{L}_{\mathrm{lat}}\), the student aligns its latent imagination with the CoT-guided teacher trajectory. Together, these two signals enable the student to internalize multimodal CoT-induced reasoning into latent space while requiring no auxiliary information at inference time.

\section{Experiments}

\subsection{Experimental Setup}

\begin{table*}[t]
\centering
\footnotesize
\renewcommand{\arraystretch}{1.08}
\setlength{\tabcolsep}{3.0pt}
\begin{tabular*}{\textwidth}{@{\extracolsep{\fill}}lccccccccccccc@{}}
\toprule
\textbf{Method}
& \multicolumn{3}{c}{\textbf{V$^\star$}}
& \multicolumn{3}{c}{\textbf{HR-4K}}
& \multicolumn{3}{c}{\textbf{HR-8K}}
& \textbf{MMStar}
& \textbf{Seed2+}
& \textbf{BLINK}
& \textbf{Hall.} \\
\cmidrule(lr){2-4}
\cmidrule(lr){5-7}
\cmidrule(lr){8-10}
& \textbf{Ovr.} & \textbf{Attr.} & \textbf{Spat.}
& \textbf{Ovr.} & \textbf{FSP} & \textbf{FCP}
& \textbf{Ovr.} & \textbf{FSP} & \textbf{FCP}
&  &  &  &  \\
\midrule

\rowcolor{groupbg}
\multicolumn{14}{l}{\textit{Proprietary Models}} \\
GPT-4o
& 67.50 & 72.20 & 60.50
& 59.00 & 70.00 & 48.00
& 55.50 & 62.00 & 49.00
& 65.20 & -- & \underline{63.00} & -- \\
Gemini-3-Flash
& \underline{86.40} & -- & --
& \underline{87.90} & -- & --
& \underline{85.00} & -- & --
& -- & -- & -- & -- \\

\midrule
\rowcolor{groupbg}
\multicolumn{14}{l}{\textit{Native Open-Source Models and SFT/OPD Variants}} \\
LLaVA-OV-7B
& 71.25 & 73.48 & 67.89
& 62.50 & 74.25 & 50.75
& 58.25 & 67.50 & 49.00
& 59.13 & 61.22 & 49.34 & 51.10 \\
Qwen-7B
& 71.20 & 73.04 & 68.42
& 65.12 & 75.75 & 54.50
& 58.00 & 63.00 & 53.00
& 59.70 & 65.31 & 53.60 & 56.57 \\
Qwen-7B + SFT
& 73.82 & 73.04 & 75.00
& 68.00 & 78.25 & 57.75
& 60.75 & 66.50 & 55.00
& 61.36 & 69.38 & 56.56 & 66.67 \\
Qwen-7B + 32B OPD
& 73.30 & 73.91 & 72.37
& 65.25 & 76.00 & 54.50
& 55.50 & 64.25 & 46.75
& 51.52 & 67.72 & 39.35 & 38.90 \\
InternVL3-8B
& 70.20 & 67.80 & 73.70
& 70.00 & 78.80 & 61.30
& 69.30 & 78.80 & \underline{59.80}
& -- & -- & -- & -- \\

\midrule
\rowcolor{groupbg}
\multicolumn{14}{l}{\textit{Thinking-with-Images Agent Models}} \\
ZoomEye
& 79.85 & 80.52 & 78.82
& 68.75 & 81.25 & 56.25
& 64.50 & 75.00 & 54.00
& 63.20 & 70.27 & 55.55 & \underline{71.08} \\
Thyme
& 82.20 & 83.50 & 80.30
& 77.00 & 91.00 & \underline{63.00}
& 72.00 & 86.50 & 57.50
& \underline{65.90} & -- & 56.10 & 55.60 \\
DeepEyes
& 83.25 & -- & --
& 75.10 & -- & --
& 72.60 & -- & --
& 58.73 & 69.08 & 51.08 & 62.57 \\
DeepEyesV2
& 81.80 & -- & --
& 77.90 & -- & --
& 73.80 & -- & --
& -- & -- & -- & -- \\

\midrule
\rowcolor{groupbg}
\multicolumn{14}{l}{\textit{Visual-Latent Models}} \\
LVR
& 80.60 & 81.70 & 79.00
& -- & -- & --
& -- & -- & --
& 57.93 & 47.39 & 53.60 & 65.19 \\
Laser
& -- & -- & --
& 72.50 & -- & --
& -- & -- & --
& 60.27 & 70.05 & 56.92 & \textbf{67.72} \\
SkiLa
& 78.53 & -- & --
& 72.12 & -- & --
& 66.50 & -- & --
& -- & -- & -- & -- \\
Monet
& 80.10 & 81.73 & 77.63
& 67.37 & 78.25 & 56.50
& 64.37 & 74.25 & 54.49
& 60.33 & 65.88 & 50.71 & 56.36 \\
HyLaR
& 83.77 & 82.61 & \textbf{\underline{85.53}}
& \textbf{75.00} & \textbf{\underline{93.75}} & 56.25
& 70.50 & \textbf{\underline{88.25}} & 52.75
& 62.00 & 70.32 & 57.14 & 63.68 \\
\textbf{OPLD}
& \textbf{85.86} & \textbf{\underline{88.70}} & 81.58
& 73.75 & 90.00 & \textbf{57.50}
& \textbf{71.37} & 87.75 & \textbf{55.00}
& \textbf{63.86} & \textbf{\underline{70.88}} & \textbf{58.56} & 65.19 \\
\bottomrule
\end{tabular*}
\caption{
Main results on multimodal reasoning benchmarks. 
All scores are percentages and higher is better. 
For visual-latent methods, the best result in each column is marked in \textbf{bold}. 
The best result among all reported methods is marked with an \underline{underline}. 
``Ovr.'' denotes the overall score. 
Qwen-7B denotes Qwen2.5-VL-7B.
}
\label{tab:main_results}
\end{table*}

\paragraph{Training and Evaluation Setup.}
For fair comparison with related methods, we use Qwen2.5-VL-7B as the default backbone for both the teacher and the student. OPLD is trained on the cleaned union of Zebra-CoT~\cite{li2025zebra} and Visual-CoT~\cite{shao2024visual}. Since some auxiliary images in multimodal CoT may contain visual cues that directly reveal the final answer, the teacher could exploit such leaked evidence as a shortcut and ignore the intended reasoning process. To prevent this teacher-side hacking issue, we filter samples with potential answer leakage and obtain 217K training examples. Unless otherwise specified, we use \(K=8\) latent slots with latent dimension \(D=2048\). The distillation stage adopts top-\(k\) forward KL with \(k=128\) and latent cosine weight \(\lambda_{\mathrm{lat}}=0.05\). We train for one epoch with learning rate \(1\times10^{-6}\) on 8 NVIDIA B200 GPUs. All evaluations are conducted with LMMS-Eval~\cite{zhang2025lmms} using greedy decoding. More details about the training data cleaning pipeline are provided in the supplementary material.
We evaluate OPLD on seven multimodal reasoning benchmarks: V$^\star$, HRBench-4K, HRBench-8K, MMStar, SeedBench2-Plus, BLINK, and HallusionBench. These benchmarks cover fine-grained visual perception, high-resolution image understanding, general multimodal reasoning, and hallucination robustness. We additionally use MME-RealWorld-Lite in the ablation study to evaluate real-world multimodal robustness.

\paragraph{Baselines.}
We compare OPLD with four groups of methods: proprietary multimodal models, native open-source MLLMs, thinking-with-images agent models, and visual-latent reasoning methods. 

\subsection{Main Results}

Table~\ref{tab:main_results} summarizes the main comparison across seven benchmarks, together with fine-grained metrics on V$^\star$ and HRBench. OPLD delivers consistent improvements over Qwen2.5-VL-7B backbone across all evaluated benchmarks. 
The gains are especially clear on perception-intensive benchmarks: OPLD improves V$^\star$ from 71.20 to 85.86, HRBench-4K from 65.12 to 73.75, and HRBench-8K from 58.00 to 71.37. 
It also improves MMStar, SeedBench2-Plus, BLINK, and HallusionBench, indicating that the proposed latent distillation framework benefits both fine-grained visual perception and general multimodal reasoning.

Compared with the SFT baseline, OPLD further improves six out of seven benchmarks. 
This suggests that the improvement is not simply caused by additional supervised training on multimodal CoT data. 
Instead, the on-policy teacher--student distillation and latent trajectory alignment provide extra reasoning-process supervision. 
For example, OPLD substantially outperforms Qwen2.5-VL-7B + SFT on V$^\star$ and HRBench-8K, where models need to identify subtle visual evidence and integrate high-resolution details before answering. 
This supports our motivation that latent states should serve as internal reasoning representations rather than only as auxiliary visual feature containers.

Compared with existing visual-latent reasoning methods, OPLD achieves the best visual-latent performance on most reported metrics. This shows that learning from CoT-induced teacher trajectories is more effective than directly aligning latent states to visual features or intermediate visual embeddings. 

The result indicates that OPLD is also competitive with thinking-with-images agent models. OPLD can internalize part of the reasoning ability induced by multimodal CoT into latent computation, reducing the need for explicit visual traces at inference time.

The fine-grained results further reveal where the improvements come from. 
On V$^\star$, OPLD improves the Attribute score from 73.04 to 88.70 over Qwen2.5-VL-7B, showing stronger fine-grained visual discrimination. 
On HRBench-4K and HRBench-8K, OPLD brings large gains on FSP, improving from 75.75 to 90.00 on HRBench-4K and from 63.00 to 87.75 on HRBench-8K. 
These improvements suggest that recursive latent reasoning helps the model organize high-resolution visual evidence before answer generation. 
Meanwhile, the gains on MMStar and BLINK show that the learned latent trajectories are not limited to high-resolution perception, but also transfer to broader multimodal reasoning tasks.

\subsection{Ablation Studies}

\begin{figure}[t]
\centering
\includegraphics[width=1.02\columnwidth]{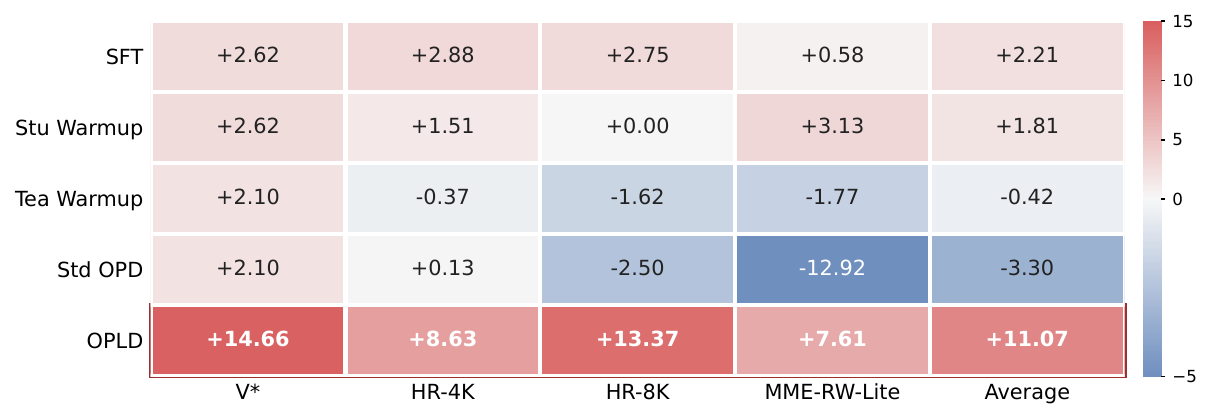}
\caption{
Training strategy ablation with the same backbone and training data.
Each cell reports the gain or drop over the Qwen2.5-VL-7B base model.
}
\label{fig:training_strategy}
\end{figure}

\paragraph{Effect of Training Strategy.}
Fig.~\ref{fig:training_strategy} compares different training strategies under the same backbone and training data, where each cell reports the performance change over the Qwen2.5-VL-7B base model. To ensure a fair comparison, all strategies use the same number of training epochs or optimizer steps, with matched batch size and learning-rate schedule. 
Standard SFT only brings limited gains, with an average improvement of \(2.21\). This suggests that simply fine-tuning in the textual CoT space is insufficient for these perception-intensive reasoning tasks, motivating the need to transfer multimodal CoT-induced reasoning into an internal latent process. Student warmup only also yields limited improvement (\(1.81\) on average). Although it introduces the latent reasoning structure, the training signal is still applied only to final answer tokens, leaving the latent slots without direct reasoning-state supervision. As a result, the student learns to use latent placeholders, but its latent imagination remains insufficient. We also find that teacher warmup only performs even worse when directly evaluated without auxiliary CoT, because the teacher is trained under a privileged input distribution with help images and multimodal rationales. Once these auxiliary inputs are removed at inference time, the model suffers from a clear distribution shift.

\begin{table}[t]
\centering
\scriptsize
\setlength{\tabcolsep}{2.6pt}
\renewcommand{\arraystretch}{1.08}
\begin{tabular}{lccccc}
\toprule
\textbf{Variant}
& \textbf{V$^\star$}
& \textbf{HR-4K}
& \textbf{HR-8K}
& \textbf{MME-RW}
& \textbf{Avg.} \\
\midrule

\multicolumn{6}{l}{\textit{Number of latent slots}}\\
\(K=0\)
&75.39&65.50&58.13&50.44&62.36\\
\(K=4\)
&78.53&72.62&\textbf{71.75}&51.90&68.70\\
\(K=8\)
&\textbf{85.86}&\textbf{73.75}&71.37&\textbf{53.36}&\textbf{71.09}\\

\midrule
\multicolumn{6}{l}{\textit{Latent architecture}}\\
Raw hidden
&78.53&71.50&67.00&48.41&66.36\\
Enc-Dec adapter
&\textbf{85.86}&\textbf{73.75}&\textbf{71.37}&\textbf{53.36}&\textbf{71.09}\\

\midrule
\multicolumn{6}{l}{\textit{Loss components}}\\
RevKL + PG + cos
&74.87&71.37&67.75&49.92&65.98\\
Top-\(k\) KL only
&78.53&63.88&58.63&48.20&62.31\\
Top-\(k\) KL + cos
&\textbf{85.86}&\textbf{73.75}&\textbf{71.37}&\textbf{53.36}&\textbf{71.09}\\

\bottomrule
\end{tabular}
\caption{
Ablation on latent design and loss components.
We report the overall accuracy on four representative benchmarks together with the average score (Avg.).
}
\label{tab:latent_loss_ablation}
\end{table}

For the standard OPD baseline, we initialize the student from Qwen2.5-VL-7B and distill it from a stronger Qwen2.5-VL-32B teacher using the same forward KL objective. 
However, this standard OPD setting is still unstable, with an average drop of \(3.30\). 
This indicates that simply transferring token-level preferences from a stronger model does not reliably improve the student's multimodal reasoning, especially when the student's latent trajectory is not explicitly aligned with the teacher's reasoning process.
In contrast, OPLD consistently improves all benchmarks. The average gain reaches \(11.07\), far exceeding all alternatives. These results validate the necessity of the full OPLD pipeline.

\paragraph{Effects of Latent Design and Loss Components.}
We further ablate three key design choices of OPLD in Table~\ref{tab:latent_loss_ablation}: the number of latent slots, the latent architecture, and the loss components. 
Increasing the number of latent slots generally improves performance. 
Compared with \(K=0\), using \(K=4\) brings clear gains on HRBench-4K and HRBench-8K, while \(K=8\) achieves the best results on V$^\star$, HRBench-4K, and MME-RealWorld-Lite. 
This suggests that multiple recursive latent steps provide useful internal computation before answer generation, and a longer latent chain gives the model stronger capacity to accumulate abstract reasoning states.

For the latent architecture, directly aligning the raw hidden space performs worse than using the encoder-decoder latent adapter. 
This supports our motivation that latent reasoning should be performed in a separate compact space rather than in the original next-token hidden space. 
For the objective design, using only the top-\(k\) token-level KL leads to a substantial drop, especially on HRBench-4K and HRBench-8K, showing that token supervision alone is insufficient to transfer the teacher's reasoning process. 
Adding cosine latent alignment largely improves the results, while the reverse-KL policy-gradient variant is less stable. 
These results confirm that stable forward KL supervision and latent trajectory alignment are complementary for effective OPLD training.

\paragraph{Latent Intervention Analysis.}
To examine whether the learned latent states functionally contribute to answer generation, we conduct controlled test-time interventions on the trained OPLD model. The results are reported in our supplementary material.

\section{Conclusion}

In this paper, we proposed \textbf{OPLD}, an on-policy latent distillation framework that transfers the abstract reasoning representations induced by privileged multimodal CoT into latent reasoning. Unlike existing methods based on feature-level latent alignment, OPLD supervises the reasoning process itself, enabling effective multimodal reasoning.
Extensive experiments show that OPLD consistently outperforms existing latent reasoning methods and achieves state-of-the-art performance on multiple benchmarks. We hope this work highlights reasoning-process supervision as a promising paradigm for learning latent reasoning in MLLMs.

\bibliography{aaai2027}


\section*{Supplementary Material}

\section{Training Data and Filtering Pipeline}

\subsection{Data Construction}

We construct the training set by merging Zebra-CoT and Visual-CoT into
a unified teacher--student format. For each sample, the student receives
only the original question and main image, while the teacher additionally
receives the textual rationale and auxiliary images:
\begin{equation}
x_S=(I,Q), \qquad
x_T=(I,Q,R,I_{\mathrm{help}}).
\end{equation}
All samples are normalized to a common schema containing the student
prompt, teacher prompt, student-visible images, teacher-visible images,
answer, sample ID, and source metadata.

\subsection{Data Cleaning}

Some auxiliary images directly reveal the final answer, which may allow
the teacher to solve the task through shortcut recognition rather than
reasoning. We therefore use the GPT-5.4 API to identify and remove such
samples.

For each sample, GPT-5.4 is given the question, answer, textual rationale,
main image, and auxiliary images. A sample is removed when the privileged
context directly displays the answer, marks the final solution, or makes
the question answerable without meaningful reasoning. Samples are retained
when the auxiliary images provide only intermediate evidence and still
require evidence integration or multi-step inference.

The exact filtering prompt is provided below.

\begin{quote}
\small
\ttfamily
You are a data-quality reviewer for multimodal reasoning datasets.

Determine whether the teacher-only multimodal context contains answer leakage.

Mark REMOVE if the auxiliary images or rationale directly reveal the final answer, such as showing the answer, marking the correct option or target, presenting the completed solution, or making meaningful reasoning unnecessary.

Mark KEEP if the context provides only intermediate evidence and meaningful reasoning is still required. When uncertain, choose KEEP.

Return only valid JSON:

\{
  ``decision'': ``KEEP'' or ``REMOVE'',
  ``reason'': ``Brief explanation.''
\}

Question: \{question\}

Ground-truth answer: \{answer\}

Rationale: \{rationale\}

Main image: [MAIN\_IMAGE]

Auxiliary images: [HELP\_IMAGES]
\end{quote}

After semantic filtering, we further remove malformed samples, invalid
image references, and duplicated IDs.

\subsection{Training Data Statistics}

Before filtering, the merged collection contains \(266{,}952\) samples:
\(178{,}695\) from Zebra-CoT and \(88{,}257\) from Visual-CoT. The
GPT-5.4-based filtering and format validation remove \(49{,}232\)
Zebra-CoT samples with potential answer leakage or invalid formatting,
reducing Zebra-CoT to \(129{,}463\) samples. The Visual-CoT split remains
unchanged. The final training set therefore contains \(217{,}720\)
samples with unique identifiers and no duplicated IDs.
Table~\ref{tab:training_data_statistics} summarizes the filtering
statistics and the main properties of the resulting dataset.

\begin{table*}[t]
\centering
\small
\setlength{\tabcolsep}{7pt}
\renewcommand{\arraystretch}{1.10}
\begin{tabular}{lrlr}
\toprule
\textbf{Statistic} & \textbf{Value}
& \textbf{Statistic} & \textbf{Value} \\
\midrule

\multicolumn{4}{l}{\textit{Filtering and dataset composition}} \\
Samples before filtering
& 266,952
& Samples after filtering
& 217,720 \\
Zebra-CoT before filtering
& 178,695
& Zebra-CoT after filtering
& 129,463 (59.5\%) \\
Visual-CoT before filtering
& 88,257
& Visual-CoT after filtering
& 88,257 (40.5\%) \\
Removed samples
& 49,232 (18.4\%)
& Duplicate sample IDs
& 0 \\

\midrule
\multicolumn{4}{l}{\textit{Teacher--student input asymmetry}} \\
Average student-visible images
& 0.97
& Average teacher-visible images
& 3.08 \\
Average auxiliary images
& 2.11
& Samples without student-visible images
& 6,580 (3.0\%) \\

\midrule
\multicolumn{4}{l}{\textit{Text length in characters: mean / median / P95}} \\
Student prompt
& 147.05 / 104 / 404
& Teacher prompt
& 1,219.34 / 783 / 3,163 \\
Answer
& 24.61 / 5 / 136
& Number of task configurations
& 13 \\

\midrule
\multicolumn{4}{l}{\textit{Task distribution in the final training set}} \\
GQA Detailed Reasoning
& 88,257 (40.5\%)
& Visual Search
& 29,393 (13.5\%) \\
Visual Jigsaw
& 21,485 (9.9\%)
& Chess
& 20,067 (9.2\%) \\
Maze
& 19,600 (9.0\%)
& Multi-Hop Object Counting
& 9,794 (4.5\%) \\
Tetris
& 9,781 (4.5\%)
& Other scientific and visual tasks
& 19,343 (8.9\%) \\

\bottomrule
\end{tabular}
\caption{
Statistics of the OPLD training data before and after filtering.
Auxiliary images denote the additional images available to the teacher
beyond the student-visible images. Text lengths are measured in
characters.
}
\label{tab:training_data_statistics}
\end{table*}

The resulting data exhibit the intended privileged teacher--student
asymmetry. The student typically receives one main image and a short
problem description, whereas the teacher observes approximately two
additional images and a substantially richer multimodal reasoning
prompt. On average, the teacher prompt is more than eight times longer
than the student prompt.

The final collection covers fine-grained visual understanding, visual
search, spatial reasoning, multi-step visual planning, and scientific
reasoning. GQA Detailed Reasoning forms the largest subset, while the
remaining tasks provide diverse reasoning trajectories involving
jigsaw puzzles, board games, mazes, object counting, and scientific
problems. The small subset without student-visible images mainly comes
from scientific reasoning tasks and is retained as complementary
abstract reasoning supervision.

\begin{algorithm}[t]
\caption{Three-Stage Training of OPLD}
\label{alg:opld}
\begin{algorithmic}[1]
\REQUIRE Training set
\(\mathcal{D}=\{(x_S,x_T,y^{\ast})\}\);
base model \(\psi_0\);
latent steps \(K\);
top-\(k\) size \(k_{\mathrm{KL}}\);
loss weights \(\lambda_{\mathrm{text}}\) and
\(\lambda_{\mathrm{lat}}\)
\ENSURE Distilled student parameters \(\theta\)

\STATE Initialize teacher \(\phi\leftarrow\psi_0\)
and student \(\theta\leftarrow\psi_0\)

\STATE \textbf{Stage 1: Teacher warmup}
\FOR{each minibatch \((x_T,y^{\ast})\) from \(\mathcal{D}\)}
    \STATE \(z^T_{1:K}\leftarrow f^{\mathrm{lat}}_{\phi}(x_T)\)
    \STATE
    \(\displaystyle
    \mathcal{L}_{T}\leftarrow
    -\sum_{n=1}^{N}
    \log p_{\phi}
    (y^{\ast}_{n}\mid x_T,z^T_{1:K},y^{\ast}_{<n})
    \)
    \STATE Update \(\phi\) using \(\nabla_{\phi}\mathcal{L}_{T}\)
\ENDFOR

\STATE \textbf{Stage 2: Student warmup}
\FOR{each minibatch \((x_S,y^{\ast})\) from \(\mathcal{D}\)}
    \STATE \(z^S_{1:K}\leftarrow f^{\mathrm{lat}}_{\theta}(x_S)\)
    \STATE
    \(\displaystyle
    \mathcal{L}_{S}\leftarrow
    -\sum_{n=1}^{N}
    \log p_{\theta}
    (y^{\ast}_{n}\mid x_S,z^S_{1:K},y^{\ast}_{<n})
    \)
    \STATE Update \(\theta\) using
    \(\nabla_{\theta}\mathcal{L}_{S}\)
\ENDFOR

\STATE Freeze teacher parameters \(\phi\)

\STATE \textbf{Stage 3: On-policy latent distillation}
\FOR{each minibatch \((x_S,x_T)\) from \(\mathcal{D}\)}

    \STATE \textit{Student on-policy rollout}
    \STATE \(z^S_{1:K}\leftarrow f^{\mathrm{lat}}_{\theta}(x_S)\)
    \STATE
    \(\hat{y}\leftarrow
    \mathrm{GreedyDecode}
    (\pi_{\theta}(\cdot\mid x_S,z^S_{1:K}))\)

    \STATE \textit{Teacher latent-target forward}
    \STATE \(z^T_{1:K}\leftarrow f^{\mathrm{lat}}_{\phi}(x_T)\)
    \STATE
    \(\displaystyle
    \mathcal{L}_{\mathrm{lat}}\leftarrow
    \frac{1}{BK}
    \sum_{b=1}^{B}\sum_{j=1}^{K}
    \left[
    1-\cos
    \left(
    z^j_{S,b},
    \mathrm{sg}(z^j_{T,b})
    \right)
    \right]
    \)

    \STATE \textit{Teacher scoring forward}
    \FOR{\(t=1,\ldots,|\hat{y}|\)}
        \STATE
        \(\displaystyle
        p^T_t\leftarrow
        p_{\phi}
        \left(
        \cdot\mid
        x_T,\mathrm{sg}(z^S_{1:K}),\hat{y}_{<t}
        \right)
        \)
        \STATE
        \(\displaystyle
        p^S_t\leftarrow
        p_{\theta}
        \left(
        \cdot\mid
        x_S,z^S_{1:K},\hat{y}_{<t}
        \right)
        \)
        \STATE
        \(\mathcal{V}^T_t\leftarrow
        \mathrm{TopK}(p^T_t,k_{\mathrm{KL}})\)
    \ENDFOR

    \STATE
    \(\displaystyle
    \mathcal{L}_{\mathrm{text}}\leftarrow
    \frac{1}{|\hat{y}|}
    \sum_{t=1}^{|\hat{y}|}
    \sum_{v\in\mathcal{V}^T_t}
    p^T_t(v)
    \left[
    \log p^T_t(v)-\log p^S_t(v)
    \right]
    \)

    \STATE
    \(\mathcal{L}_{\mathrm{OPLD}}\leftarrow
    \lambda_{\mathrm{text}}\mathcal{L}_{\mathrm{text}}
    +
    \lambda_{\mathrm{lat}}\mathcal{L}_{\mathrm{lat}}\)

    \STATE Update \(\theta\) using
    \(\nabla_{\theta}\mathcal{L}_{\mathrm{OPLD}}\)

\ENDFOR

\RETURN \(\theta\)
\end{algorithmic}
\end{algorithm}

\section{Additional Implementation Details}

Both the teacher and student are initialized from Qwen2.5-VL-7B-Instruct and adopt the same recursive latent architecture. We insert \(K=8\) latent slots with latent dimension \(D=2048\) between the assistant generation prompt and the answer sequence. The latent states are generated in the \texttt{continuous\_feedback} mode, where each decoded latent embedding is written back into the corresponding slot before computing the next latent state. The teacher and student are independently warmed up with privileged and original inputs, respectively, and the warmed-up checkpoints are then used for OPLD distillation. Table~\ref{tab:implementation_details} summarizes the main training configurations.

\begin{table*}[t]
\centering
\small
\setlength{\tabcolsep}{5.5pt}
\renewcommand{\arraystretch}{1.10}
\begin{tabular}{lccc}
\toprule
\textbf{Configuration}
& \textbf{Teacher Warmup}
& \textbf{Student Warmup}
& \textbf{OPLD Distillation} \\
\midrule
Initialization
& Qwen2.5-VL-7B-Instruct
& Qwen2.5-VL-7B-Instruct
& Warmed-up student/teacher \\

Input
& Image, question, multimodal CoT
& Image and question
& Student and teacher views \\

Trainable model
& Full model
& Full model
& Student only \\

Training objective
& Answer CE
& Answer CE
& Top-\(k\) forward KL + latent cosine \\

Answer-loss weight
& \(1.0\)
& \(1.0\)
& -- \\

Latent-loss weight
& \(0\)
& \(0\)
& \(0.05\) \\

Top-\(k\)
& --
& --
& \(128\) \\

Number of latent slots
& \(8\)
& \(8\)
& \(8\) \\

Latent dimension
& \(2048\)
& \(2048\)
& \(2048\) \\

Learning rate
& \(1\times10^{-5}\)
& \(1\times10^{-5}\)
& \(1\times10^{-6}\) \\

Vision encoder learning rate
& \(2\times10^{-6}\)
& \(2\times10^{-6}\)
& -- \\

Weight decay
& \(0.1\)
& \(0.1\)
& -- \\

LR scheduler
& Cosine
& Cosine
& -- \\

Warmup ratio
& \(0.03\)
& \(0.03\)
& -- \\

Global batch size
& \(16\)
& \(16\)
& -- \\

Training epochs
& \(1\)
& \(1\)
& \(1\) \\

Maximum prompt length
& --
& --
& \(8192\) \\

Maximum generation length
& \(512\)
& \(512\)
& \(256\) \\

Decoding strategy
& Greedy
& Greedy
& Greedy \\

Precision
& BF16
& BF16
& BF16 \\

Attention implementation
& SDPA
& SDPA
& -- \\

Gradient checkpointing
& Enabled
& Enabled
& Enabled \\

Teacher correction mode
& --
& --
& Free rollout \\

Policy gradient / task reward
& --
& --
& Disabled \\

\bottomrule
\end{tabular}
\caption{
Training configurations for the three stages of OPLD. The teacher and
student warmup stages use answer-only supervised learning, while the
distillation stage combines token-level forward KL and latent trajectory
alignment.
}
\label{tab:implementation_details}
\end{table*}

During warmup, we perform full-parameter fine-tuning. The language model and multimodal merger use a learning rate of \(1\times10^{-5}\), while the vision encoder uses \(2\times10^{-6}\). Training uses BF16 precision, TF32 computation, scaled dot-product attention, the Liger kernel, and non-reentrant gradient checkpointing. The per-device batch size is \(1\), and gradient accumulation is used to obtain a global batch size of \(16\). Input images are dynamically resized within a pixel range of \(100{,}352\) to \(1{,}317{,}120\). The answer tokens are supervised with cross-entropy, while prompt tokens and latent slots are excluded from the loss.

In the distillation stage, the teacher is frozen and only the student is updated. The student first generates an on-policy latent trajectory and response using greedy decoding. The teacher then independently produces its privileged latent trajectory for cosine alignment and scores the student-generated trajectory for token-level distillation. We use forward KL over the \(128\) highest-probability teacher tokens and set the latent cosine weight to \(0.05\). Policy-gradient optimization and task-specific rewards are disabled.
All reported evaluations are conducted with LMMS-Eval using greedy decoding and the official task configurations.

\section{Additional Method Details}

\subsection{Training Algorithm}

Algorithm~\ref{alg:opld} summarizes the complete three-stage training
procedure. The teacher and student are initialized from the same
pretrained MLLM but are optimized independently during warmup. In the
final stage, the teacher is frozen and only the student is updated.

\subsection{Discussion}

\paragraph{Warmup and latent-space compatibility.}
The two warmup stages establish compatible but not necessarily identical
latent reasoning spaces. The teacher is warmed up with privileged
multimodal CoT, enabling its recursive latent trajectory to exploit
textual rationales and auxiliary images. The student is independently
warmed up using only the original image and question, which prevents its
latent slots from remaining unstructured before distillation.

Since both models undergo full-parameter warmup, their latent
representations are not mathematically guaranteed to remain in exactly
the same coordinate system. OPLD therefore does not assume that
\(z_k^S\) and \(z_k^T\) are already coordinate-wise equivalent before
distillation. Instead, the initial mismatch is reduced by shared
structural priors: the teacher and student are initialized from the same
pretrained MLLM, employ latent encoder--decoder modules with the same
architecture and initialization, use the same latent dimensionality, and
follow the same recursive slot order. Their latent spaces thus originate
from a common representation basis, although they may drift during
independent warmup.

The frozen teacher latent space subsequently acts as the reference space
during distillation. The slot-wise cosine objective is therefore a
calibration objective rather than an assumption of pre-existing latent
equivalence:
\[
\mathcal{L}_{\mathrm{lat}}
=
\frac{1}{K}
\sum_{k=1}^{K}
\left(
1-
\frac{
\langle z_k^S,z_k^T\rangle
}{
\lVert z_k^S\rVert_2
\lVert z_k^T\rVert_2
}
\right).
\]
All student parameters, including its latent encoder and decoder, remain
trainable, allowing the student representation to be progressively
calibrated toward the frozen teacher space. Cosine distance further
reduces sensitivity to differences in latent magnitude.

\paragraph{Tensor-level path of the three forward processes.}
For clarity, we distinguish the student rollout, teacher latent-target
rollout, and teacher scoring forward. Let
\(E_\theta,D_\theta\) denote the student latent encoder and decoder, and
let \(E_\phi,D_\phi\) denote their frozen teacher counterparts.

During the student rollout, the student recursively generates
\[
z_k^S
=
E_\theta\!\left(h^{S,L}_{p_k-1}\right),
\qquad
e_k^S
=
D_\theta(z_k^S),
\]
where \(h^{S,L}_{p_k-1}\) is the last-layer hidden state immediately
preceding the \(k\)-th latent slot. The decoded embedding \(e_k^S\) is
written into the \(k\)-th student slot before the next latent state is
computed. After \(K\) recursive steps, the student generates the answer
trajectory \(\hat y\) using its current parameters. Although greedy
decoding is deterministic, this trajectory is on-policy because it is
generated by the current student rather than from ground-truth prefixes
or an offline teacher trajectory.

The teacher latent-target rollout independently performs the same
recursive computation under the privileged input \(x_T\):
\[
z_k^T
=
E_\phi\!\left(h^{T,L}_{p_k-1}\right),
\qquad
e_k^T
=
D_\phi(z_k^T).
\]
This pass invokes both the teacher encoder and teacher decoder and
produces the privileged trajectory \(z^T_{1:K}\) used by
\(\mathcal{L}_{\mathrm{lat}}\).

The teacher scoring forward follows a different path. It does not
recompute teacher latent codes and does not reuse the student-decoded
embeddings \(e_k^S\). Instead, each stop-gradient student code is decoded
by the frozen teacher decoder:
\[
\widetilde e_k^{S\rightarrow T}
=
D_\phi\!\left(
\operatorname{sg}(z_k^S)
\right),
\qquad k=1,\ldots,K,
\]
where \(\operatorname{sg}(\cdot)\) denotes stop-gradient. The resulting
embeddings are inserted into the corresponding latent positions of the
teacher input sequence:
\[
\mathbf{E}^{T,\mathrm{score}}
=
\Big[
\operatorname{Emb}_\phi(x_T);
\widetilde e_1^{S\rightarrow T},
\ldots,
\widetilde e_K^{S\rightarrow T};
\operatorname{Emb}_\phi(\hat y_{<t})
\Big].
\]
The teacher then computes
\[
p_t^T
=
p_\phi
\left(
\cdot
\mid
x_T,
\operatorname{sg}(z^S_{1:K}),
\hat y_{<t}
\right).
\]

Thus, the teacher latent encoder \(E_\phi\) is bypassed in the scoring
forward because the latent codes are supplied by the student. The
teacher decoder \(D_\phi\) is used to interpret these codes in the
teacher's embedding interface. This differs from directly inserting
\(e_k^S=D_\theta(z_k^S)\), which would condition the teacher on
student-decoded embeddings rather than on student latent codes and would
not match the notation above.

A tensor-level summary is:
\[
\begin{aligned}
\text{Student rollout:}\quad
&h^S
\xrightarrow{E_\theta}
z^S
\xrightarrow{D_\theta}
e^S
\xrightarrow{\text{student MLLM}}
\hat y,\\
\text{Teacher target:}\quad
&h^T
\xrightarrow{E_\phi}
z^T
\xrightarrow{D_\phi}
e^T,\\
\text{Teacher scoring:}\quad
&\operatorname{sg}(z^S)
\xrightarrow{D_\phi}
\widetilde e^{S\rightarrow T}
\xrightarrow{\text{teacher MLLM}}
p^T.
\end{aligned}
\]

The teacher parameters are frozen in both teacher forwards. Teacher
outputs and teacher-decoded student embeddings are treated as detached
targets; gradients are propagated only through the student computation.

\paragraph{Why decode student codes with the teacher decoder?}
Using \(D_\phi\) ensures that the scoring forward receives embeddings
expressed through the teacher's own latent-to-embedding interface. At the
beginning of distillation, the student codes may still be imperfectly
calibrated for \(D_\phi\). However, the shared initialization and student
warmup prevent them from being arbitrary, while
\(\mathcal{L}_{\mathrm{lat}}\) progressively moves \(z_k^S\) toward the
region represented by \(z_k^T\). The teacher scoring signal therefore
becomes increasingly reliable as latent calibration improves.

The encoder--decoder bottleneck is useful in this setting because it
provides an explicit, trainable interface for correcting
teacher--student representation drift. This interpretation is consistent
with the empirical advantage of aligning compact encoder--decoder
latents over directly matching the original high-dimensional MLLM hidden
states.

\paragraph{Complementary distillation objectives.}
The teacher scoring forward evaluates token preferences at the latent
and textual states visited by the current student. The top-\(k\) forward
KL objective transfers these preferences:
\[
\mathcal{L}_{\mathrm{text}}
=
\frac{1}{T}
\sum_{t=1}^{T}
D_{\mathrm{KL}}
\left(
p_{t,\mathrm{top}\text{-}k}^{T}
\;\middle\|\;
p_{t,\mathrm{top}\text{-}k}^{S}
\right).
\]
However, token-level supervision alone does not explicitly calibrate the
student latent coordinates. Conversely, latent cosine alignment does not
guarantee that the aligned states induce the desired answer
distribution. OPLD therefore combines the two objectives:
\[
\mathcal{L}_{\mathrm{OPLD}}
=
\mathcal{L}_{\mathrm{text}}
+
\lambda_{\mathrm{lat}}
\mathcal{L}_{\mathrm{lat}}.
\]
No policy-gradient loss or task-specific reward is used in the final
configuration.

\paragraph{Interpretation and limitation of slot-wise alignment.}
The fixed slot index provides a shared structural correspondence: the
\(k\)-th state in both models is computed after \(k-1\) recursive latent
feedback steps. It therefore represents the same recursive depth and
removes the permutation ambiguity associated with an unordered set of
latent tokens. Nevertheless, equal recursive depth does not guarantee
that the teacher and student encode exactly the same semantic reasoning
stage at slot \(k\). The slot-wise cosine loss should consequently be
viewed as an effective temporal inductive bias, rather than a theoretical
guarantee of one-to-one semantic correspondence.

Our ablations show that this ordered alignment improves downstream
performance, while the latent intervention experiments demonstrate that
the input-conditioned latent trajectory functionally affects answer
generation. These results support the practical utility of the learned
trajectory, but they do not establish an interpretable semantic meaning
for every slot or dimension. More flexible cross-space alignment, such
as a learned latent mapper, optimal-transport matching, or diagnostic
probing of individual slots, remains an important direction for future
work.

\paragraph{Inference.}
At inference time, the teacher, teacher decoder, and all privileged
inputs are removed. The student receives only the original image and
question, performs \(K=8\) recursive latent steps using
\(E_\theta\) and \(D_\theta\), and then generates the answer. OPLD
therefore requires neither textual CoT generation nor auxiliary visual
operations during deployment.

\section{Additional Experimental Results}

\subsection{Fine-Grained Training Strategy Comparison}

Table~\ref{tab:supp_training_strategy} reports the complete results of
different training strategies, including the fine-grained metrics of
V$^\star$, HRBench, and MME-RealWorld-Lite. All variants use the same
Qwen2.5-VL-7B student backbone and training data.

\begin{table*}[t]
\centering
\footnotesize
\setlength{\tabcolsep}{2.6pt}
\renewcommand{\arraystretch}{1.08}
\begin{tabular*}{\textwidth}{@{\extracolsep{\fill}}lcccccccccccc@{}}
\toprule
\textbf{Method}
& \multicolumn{3}{c}{\textbf{V$^\star$}}
& \multicolumn{3}{c}{\textbf{HRBench-4K}}
& \multicolumn{3}{c}{\textbf{HRBench-8K}}
& \multicolumn{3}{c}{\textbf{MME-RW-Lite}} \\
\cmidrule(lr){2-4}
\cmidrule(lr){5-7}
\cmidrule(lr){8-10}
\cmidrule(lr){11-13}
& \textbf{Ovr.} & \textbf{Attr.} & \textbf{Spat.}
& \textbf{Ovr.} & \textbf{FSP} & \textbf{FCP}
& \textbf{Ovr.} & \textbf{FSP} & \textbf{FCP}
& \textbf{Ovr.} & \textbf{Reason.} & \textbf{Percep.} \\
\midrule
Qwen2.5-VL-7B
& 71.20 & 73.04 & 68.42
& 65.12 & 75.75 & 54.50
& 58.00 & 63.00 & 53.00
& 45.75 & 39.73 & 49.62 \\

Qwen2.5-VL-7B + SFT
& 73.82 & 73.04 & 75.00
& 68.00 & 78.25 & 57.75
& 60.75 & 66.50 & \textbf{55.00}
& 46.33 & 42.40 & 48.85 \\

Student Warmup
& 73.82 & 72.17 & 76.32
& 66.63 & 73.50 & \textbf{59.75}
& 58.00 & 61.50 & 54.50
& 48.88 & 44.53 & 51.67 \\

Teacher Warmup
& 73.30 & 73.04 & 73.68
& 64.75 & 72.25 & 57.25
& 56.38 & 61.25 & 51.50
& 43.98 & 38.53 & 47.48 \\

Qwen2.5-VL-7B + 32B OPD
& 73.30 & 73.91 & 72.37
& 65.25 & 76.00 & 54.50
& 55.50 & 64.25 & 46.75
& 32.83 & 25.60 & 37.47 \\

\textbf{OPLD}
& \textbf{85.86} & \textbf{88.70} & \textbf{81.58}
& \textbf{73.75} & \textbf{90.00} & 57.50
& \textbf{71.37} & \textbf{87.75} & \textbf{55.00}
& \textbf{53.36} & \textbf{47.87} & \textbf{56.89} \\
\bottomrule
\end{tabular*}
\caption{
Fine-grained comparison of different training strategies. ``Ovr.'',
``Attr.'', ``Spat.'', ``Reason.'', and ``Percep.'' denote overall,
attribute, spatial, reasoning, and perception scores, respectively.
The best result in each column is highlighted in bold.
}
\label{tab:supp_training_strategy}
\end{table*}

OPLD provides the most consistent improvements across both overall and
fine-grained metrics. Compared with the base model, it improves the
V$^\star$ Attribute and Spatial scores by \(15.66\) and \(13.16\)
points, respectively. On HRBench, the largest gains occur on FSP,
increasing from \(75.75\) to \(90.00\) at 4K resolution and from
\(63.00\) to \(87.75\) at 8K resolution. These results indicate that
OPLD substantially strengthens fine-grained evidence perception and
high-resolution visual reasoning. It also improves both the Reasoning
and Perception components of MME-RealWorld-Lite.

In contrast, ordinary SFT and student warmup provide only moderate and
inconsistent gains. Teacher warmup degrades when privileged CoT is
removed during evaluation, reflecting the input-distribution shift
between training and inference. Standard OPD with a larger 32B teacher
also performs poorly on HRBench-8K and MME-RealWorld-Lite, showing that
token-level distillation from a stronger teacher alone does not
reliably transfer multimodal reasoning ability.

\subsection{Fine-Grained Ablation Results}

Table~\ref{tab:supp_fine_ablation} presents the complete fine-grained
results for the latent-slot number, latent architecture, and
distillation objectives.

\begin{table*}[t]
\centering
\footnotesize
\setlength{\tabcolsep}{2.4pt}
\renewcommand{\arraystretch}{1.08}
\begin{tabular*}{\textwidth}{@{\extracolsep{\fill}}lcccccccccccc@{}}
\toprule
\textbf{Variant}
& \multicolumn{3}{c}{\textbf{V$^\star$}}
& \multicolumn{3}{c}{\textbf{HRBench-4K}}
& \multicolumn{3}{c}{\textbf{HRBench-8K}}
& \multicolumn{3}{c}{\textbf{MME-RW-Lite}} \\
\cmidrule(lr){2-4}
\cmidrule(lr){5-7}
\cmidrule(lr){8-10}
\cmidrule(lr){11-13}
& \textbf{Ovr.} & \textbf{Attr.} & \textbf{Spat.}
& \textbf{Ovr.} & \textbf{FSP} & \textbf{FCP}
& \textbf{Ovr.} & \textbf{FSP} & \textbf{FCP}
& \textbf{Ovr.} & \textbf{Reason.} & \textbf{Percep.} \\
\midrule

\multicolumn{13}{l}{\textit{Number of latent slots}} \\
\(K=0\)
& 75.39 & 75.65 & 75.00
& 65.50 & 74.25 & 56.75
& 58.13 & 61.75 & 54.50
& 50.44 & 46.27 & 53.12 \\

\(K=4\)
& 78.53 & 77.39 & 80.26
& 72.62 & 89.25 & 56.00
& \textbf{71.75} & 86.00 & \textbf{57.50}
& 51.90 & 45.87 & 55.77 \\

\(K=8\)
& \textbf{85.86} & \textbf{88.70} & \textbf{81.58}
& \textbf{73.75} & \textbf{90.00} & \textbf{57.50}
& 71.37 & \textbf{87.75} & 55.00
& \textbf{53.36} & \textbf{47.87} & \textbf{56.89} \\

\midrule
\multicolumn{13}{l}{\textit{Latent architecture}} \\
Raw hidden states
& 78.53 & 80.87 & 75.00
& 71.50 & 88.25 & 54.75
& 67.00 & 79.25 & 54.75
& 48.41 & 43.73 & 51.41 \\

Enc--Dec adapter
& \textbf{85.86} & \textbf{88.70} & \textbf{81.58}
& \textbf{73.75} & \textbf{90.00} & \textbf{57.50}
& \textbf{71.37} & \textbf{87.75} & \textbf{55.00}
& \textbf{53.36} & \textbf{47.87} & \textbf{56.89} \\

\midrule
\multicolumn{13}{l}{\textit{Distillation objectives}} \\
Reverse KL + PG + cosine
& 74.87 & 76.52 & 72.37
& 71.37 & 85.75 & 57.00
& 67.75 & 81.50 & 54.00
& 49.92 & 43.33 & 54.15 \\

Top-\(k\) forward KL only
& 78.53 & 80.87 & 75.00
& 63.88 & 74.75 & 53.00
& 58.63 & 66.25 & 51.00
& 48.20 & 43.73 & 51.07 \\

Top-\(k\) forward KL + cosine
& \textbf{85.86} & \textbf{88.70} & \textbf{81.58}
& \textbf{73.75} & \textbf{90.00} & \textbf{57.50}
& \textbf{71.37} & \textbf{87.75} & \textbf{55.00}
& \textbf{53.36} & \textbf{47.87} & \textbf{56.89} \\

\bottomrule
\end{tabular*}
\caption{
Fine-grained ablation results for the number of latent slots, latent
architecture, and distillation objectives. The best result within each
ablation group is highlighted in bold.
}
\label{tab:supp_fine_ablation}
\end{table*}

Increasing the number of latent slots generally improves performance.
Although \(K=4\) slightly outperforms \(K=8\) on the HRBench-8K overall
and FCP metrics, \(K=8\) achieves the strongest results on most other
metrics, especially V$^\star$ Attribute and HRBench FSP. This suggests
that additional recursive latent steps primarily benefit the
integration of fine-grained and spatially distributed visual evidence.

The encoder--decoder adapter consistently outperforms direct alignment
in the raw hidden-state space. The largest improvements appear on
V$^\star$ and HRBench-8K FSP, supporting the use of a compact projected
space for learning abstract reasoning representations.

Finally, top-\(k\) forward KL alone performs substantially worse on
HRBench, despite achieving reasonable V$^\star$ results. Adding cosine
latent alignment improves every reported metric, demonstrating that
token-level feedback alone is insufficient to transfer the teacher's
intermediate reasoning process. The reverse-KL policy-gradient variant
is also less effective than direct top-\(k\) forward KL combined with
latent trajectory alignment.

\subsection{Latent Intervention Analysis}

To evaluate whether the learned latent trajectory functionally
contributes to answer generation, we conduct controlled test-time
interventions on the same trained OPLD model. We keep the model
parameters, original input, and decoding configuration unchanged, and
modify only the latent states used before answer generation. We compare
the original trajectory with three interventions: adding random noise,
setting all latent states to zero, and shuffling latent trajectories
across different samples. No additional training is performed for any
intervention.

\begin{table}[t]
\centering
\scriptsize
\setlength{\tabcolsep}{3.0pt}
\renewcommand{\arraystretch}{1.08}
\begin{tabular}{lccccc}
\toprule
\textbf{Latent Setting}
& \textbf{V$^\star$}
& \textbf{HR-4K}
& \textbf{HR-8K}
& \textbf{MME-RW}
& \textbf{Avg. Drop} \\
\midrule
Original Latents
& \textbf{85.86}
& \textbf{73.75}
& \textbf{71.37}
& \textbf{53.36}
& -- \\

Noisy Latents
& 76.96
& 64.12
& 57.50
& 48.62
& 9.29 \\

Zeroed Latents
& 76.44
& 59.62
& 51.37
& 48.51
& 12.10 \\

Shuffled Latents
& 65.45
& 54.63
& 48.25
& 44.61
& 17.85 \\
\bottomrule
\end{tabular}
\caption{
Test-time interventions on the latent trajectory of the same trained
OPLD model. 
}
\label{tab:latent_intervention}
\end{table}

As shown in Table~\ref{tab:latent_intervention}, modifying the latent
trajectory consistently degrades performance across all four
benchmarks. Adding noise results in an average drop of \(9.29\) points,
while completely removing the latent information by zeroing the states
increases the drop to \(12.10\) points. These results show that the
answer decoder relies on information carried by the learned latent
trajectory rather than merely on the presence of latent positions.

Shuffling produces the largest degradation, reducing the average score
by \(17.85\) points. Unlike zeroing or random perturbation, shuffled
latents remain valid representations produced by the same model, but
they no longer correspond to the current input. The substantially larger
drop therefore indicates that the effectiveness of latent reasoning
depends on its sample-specific content, rather than only its numerical
scale or distribution.

The fine-grained results exhibit the same pattern. Shuffling reduces the
V$^\star$ Attribute score from \(88.70\) to \(57.39\), HRBench-4K FSP
from \(90.00\) to \(58.75\), and HRBench-8K FSP from \(87.75\) to
\(50.25\). The degradation is particularly pronounced on metrics
requiring fine-grained evidence identification and high-resolution
perception, suggesting that the recursive latent trajectory plays an
important role in integrating task-relevant visual information before
answer generation.

\subsection{Privileged Teacher Reference}

We further evaluate how closely the OPLD student approaches a teacher
that retains access to privileged multimodal CoT. The comparison is
conducted on a held-out in-domain test set containing about 5K
samples. The warmed-up teacher receives the original input together
with textual rationales and auxiliary images, whereas both student
models receive only the original question and main image.

\begin{figure}[t]
\centering
\includegraphics[width=0.95\columnwidth]
{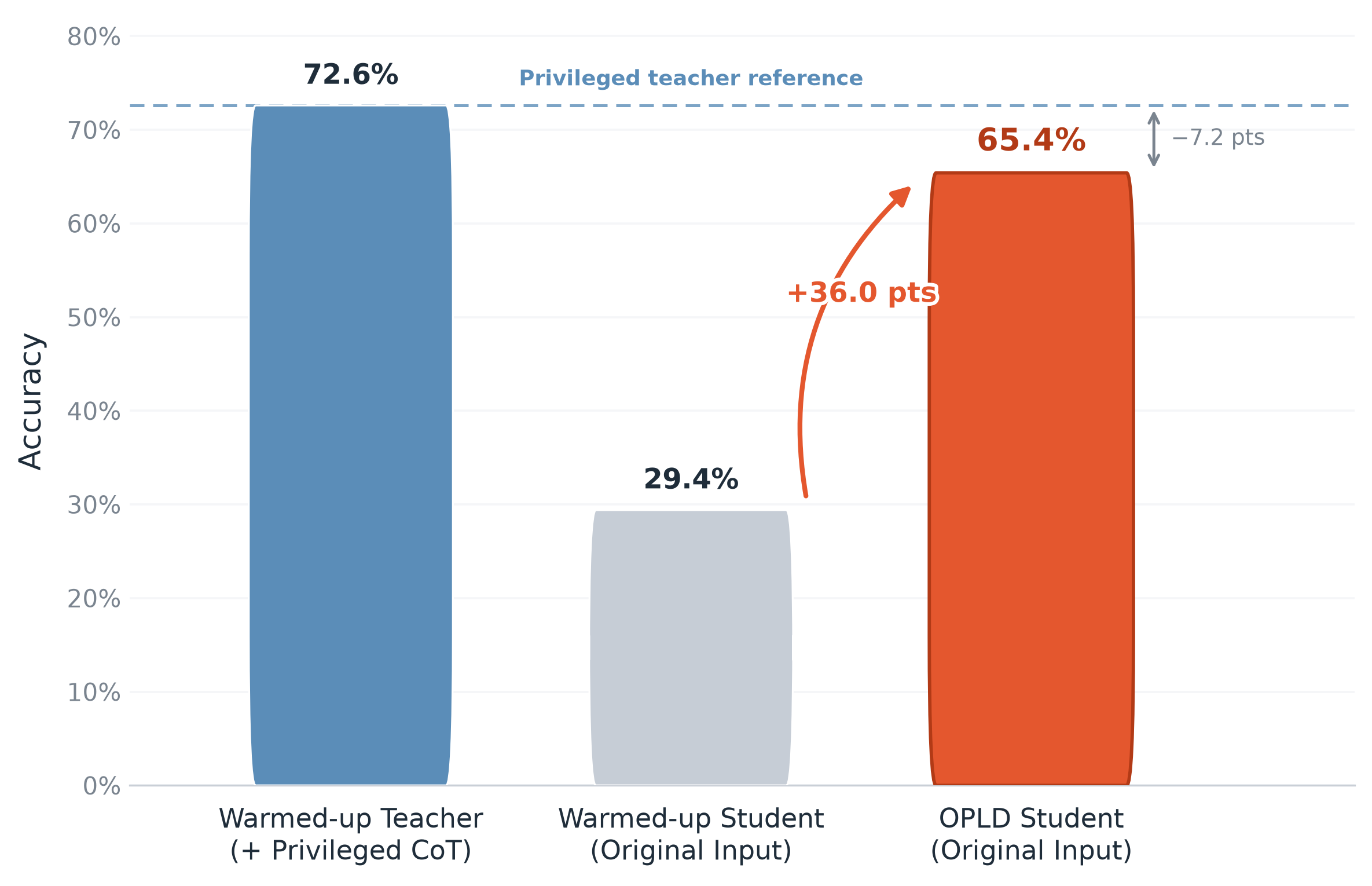}
\caption{
In-domain accuracy on 5K held-out samples. OPLD substantially closes
the gap to the teacher with privileged multimodal CoT.
}
\label{fig:privileged_teacher_gap}
\end{figure}

As shown in Fig.~\ref{fig:privileged_teacher_gap}, the warmed-up teacher
achieves \(72.6\%\) accuracy with privileged multimodal CoT, while the
warmed-up student reaches only \(29.4\%\) using the original input.
After on-policy latent distillation, the OPLD student improves to
\(65.4\%\), corresponding to a gain of \(36.0\) percentage points over
student warmup.

OPLD therefore recovers \(83.3\%\) of the initial performance gap
between the warmed-up student and the privileged teacher, leaving only
a \(7.2\)-point difference. Importantly, the OPLD student obtains this
performance without access to textual CoT or auxiliary images at
inference time. These results indicate that OPLD transfers a substantial
portion of the reasoning capability induced by privileged multimodal
CoT into the student's latent reasoning process. The teacher result is
used as a privileged reference rather than a strict theoretical upper
bound.

\section{Case Studies}

We provide a collection of qualitative case studies to compare model
predictions across the three stages of OPLD training. Each case contains
the original question, the main input image, the privileged multimodal
CoT available to the teacher, and the predictions produced by the
warmed-up teacher, the warmed-up student, and the final OPLD student.

The \textbf{Question} and \textbf{Question Image} fields correspond to
the original student input and are available to all models. The
\textbf{Privileged Multimodal CoT} field contains the textual rationale
and auxiliary images used only by the teacher during training. The
\textbf{Predictions} field reports results after the three training
stages. \textbf{Warmed-up Teacher} denotes the teacher trained with the
original input and privileged multimodal CoT. \textbf{Warmed-up Student}
denotes the student trained with answer-only supervision using the
original input. \textbf{OPLD Student} denotes the same student after
on-policy latent distillation. The warmed-up student and the OPLD student
receive identical inputs during inference, so their prediction
difference reflects the effect of the distillation stage.

For readability, the \(K\) recursive \texttt{<latent>} slots are omitted.
In the actual model input, these latent slots are inserted between the
assistant generation prompt and the answer sequence. Each model first
performs \(K\)-step recursive latent reasoning and then generates the
reported prediction. The case-study figures show only the observable
inputs and final outputs, rather than the intermediate continuous latent
states.

An interesting phenomenon is that, in several examples, the warmed-up
teacher produces an incorrect final answer, whereas the OPLD student
produces the correct answer after distillation. This does not imply that
the student directly copies and then surpasses the teacher's displayed
hard prediction. OPLD does not use the teacher's decoded answer as a
hard supervision target. Instead, it transfers two forms of soft
supervision: the teacher's token-level probability distribution on the
student-generated trajectory and the teacher's privileged latent
trajectory. A teacher may select an incorrect token under greedy
decoding while still assigning informative probability mass to the
correct token or providing useful preferences over competing answers.
Forward KL can transfer this richer distributional information without
forcing the student to reproduce the teacher's top-1 prediction.

Moreover, the displayed warmed-up teacher prediction is generated from
the teacher's own free rollout, whereas token-level distillation uses a
separate teacher scoring forward pass. In the scoring pass, the teacher
conditions on the student's latent codes and student-generated answer
prefix:
\[
p^T_t =
p_{\phi}\!\left(
\cdot \mid x_T, z^S_{1:K}, \hat{y}_{<t}
\right).
\]
Consequently, the supervision received by the student is not identical
to the sequence obtained from the teacher's independent greedy rollout.
The teacher may make an incorrect final prediction along its own
trajectory while still providing useful corrective preferences at the
states visited by the student.

The latent objective should likewise not be interpreted as requiring the
student to reproduce the teacher's complete answer behavior. It aligns
intermediate representations formed under privileged multimodal
evidence, while the student retains its own pretrained knowledge,
student-warmup initialization, and inference-time input pathway. The
final student is therefore constrained jointly by its original
capabilities, the token-level teacher signal, and the privileged latent
regularization. Because these signals are optimized over the entire
training distribution rather than on a single example, the student can
integrate transferable reasoning patterns while avoiding some
example-specific errors of the teacher.




\begin{figure*}[t]
\centering
\includegraphics[width=0.65\textwidth]{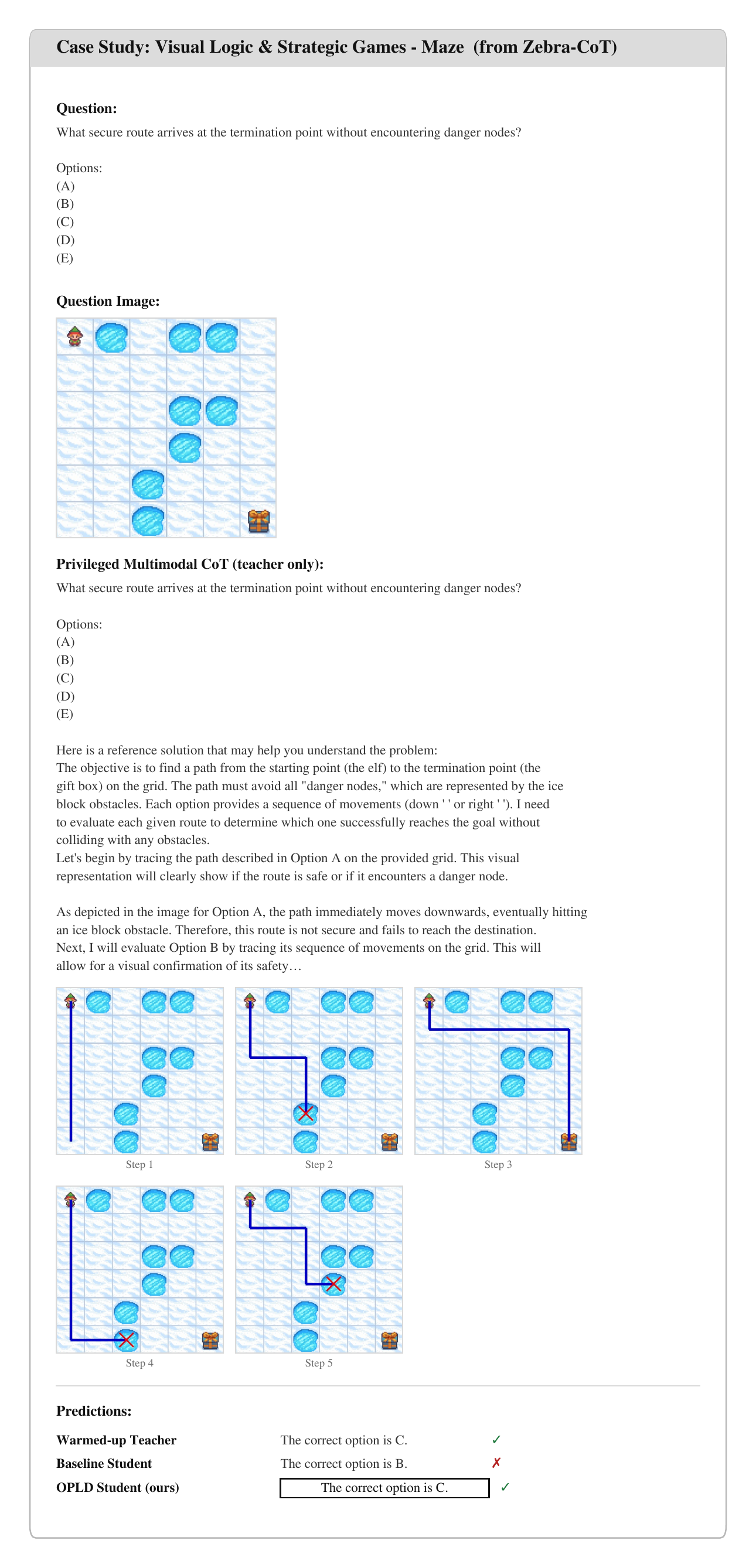}
\caption{
Inference Result
}
\end{figure*}

\begin{figure*}[t]
\centering
\includegraphics[width=0.72\textwidth]{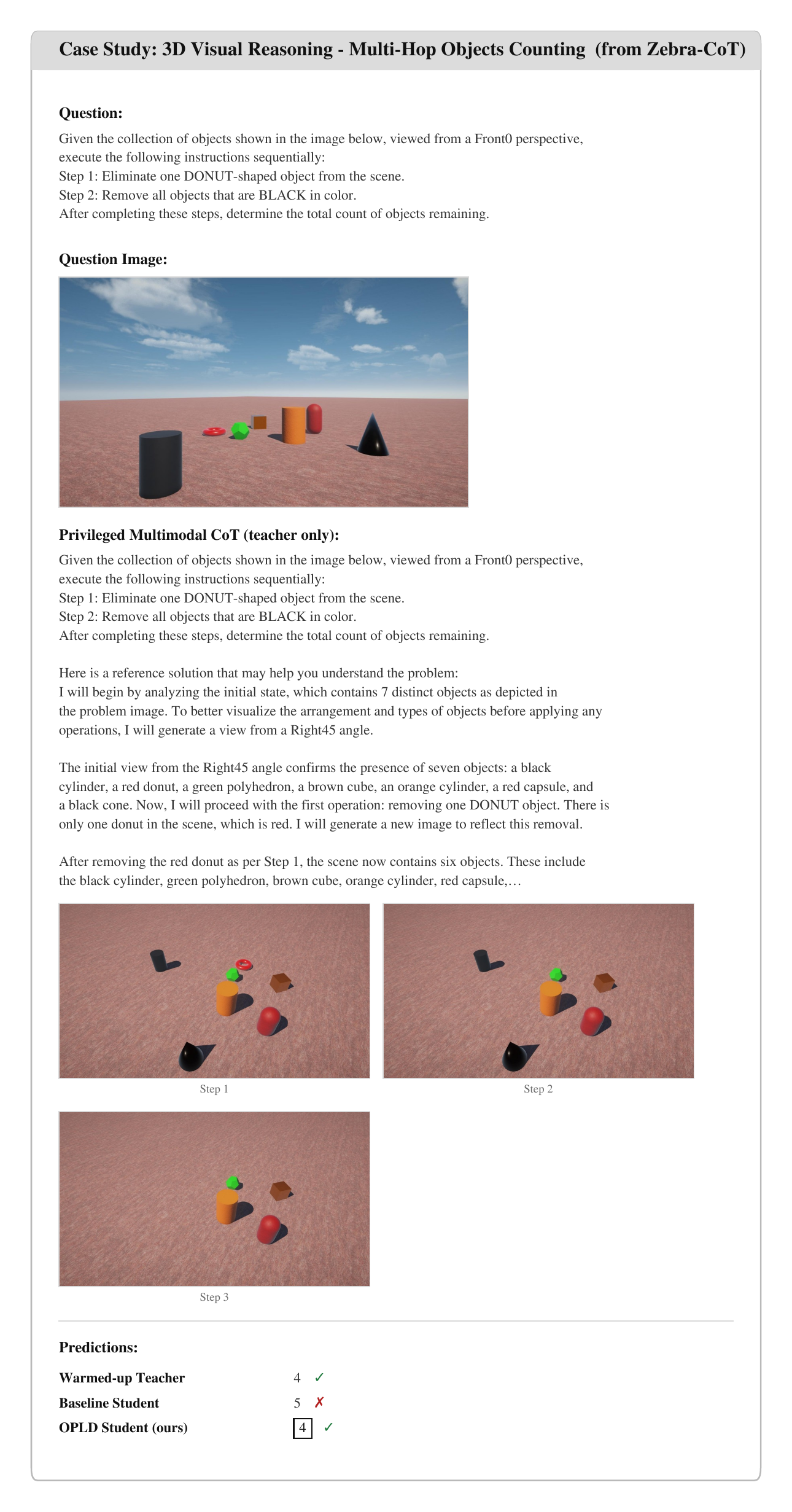}
\caption{
Inference Result
}
\end{figure*}

\begin{figure*}[t]
\centering
\includegraphics[width=0.7\textwidth]{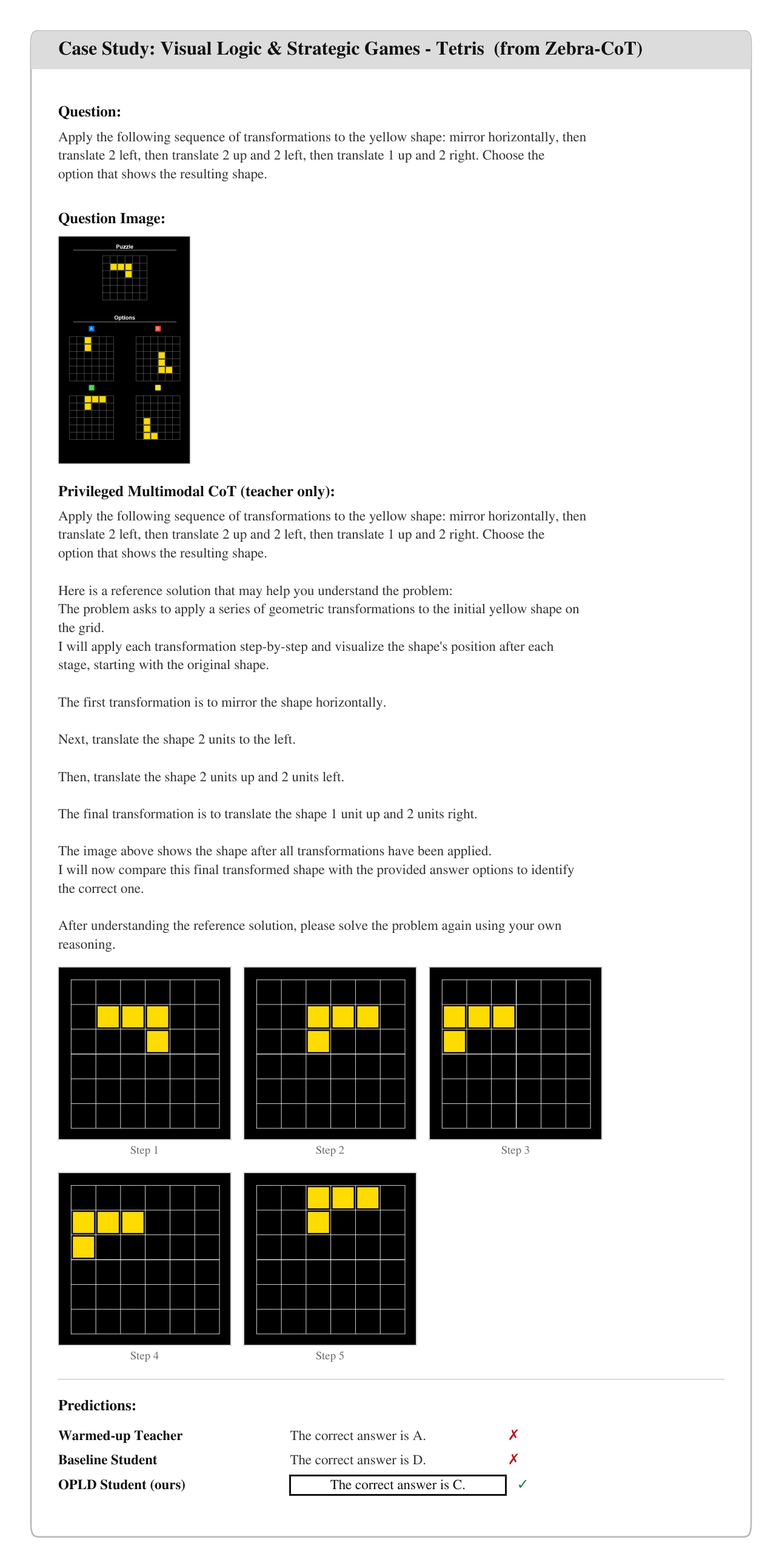}
\caption{
Inference Result
}
\end{figure*}

\begin{figure*}[t]
\centering
\includegraphics[width=0.9\textwidth]{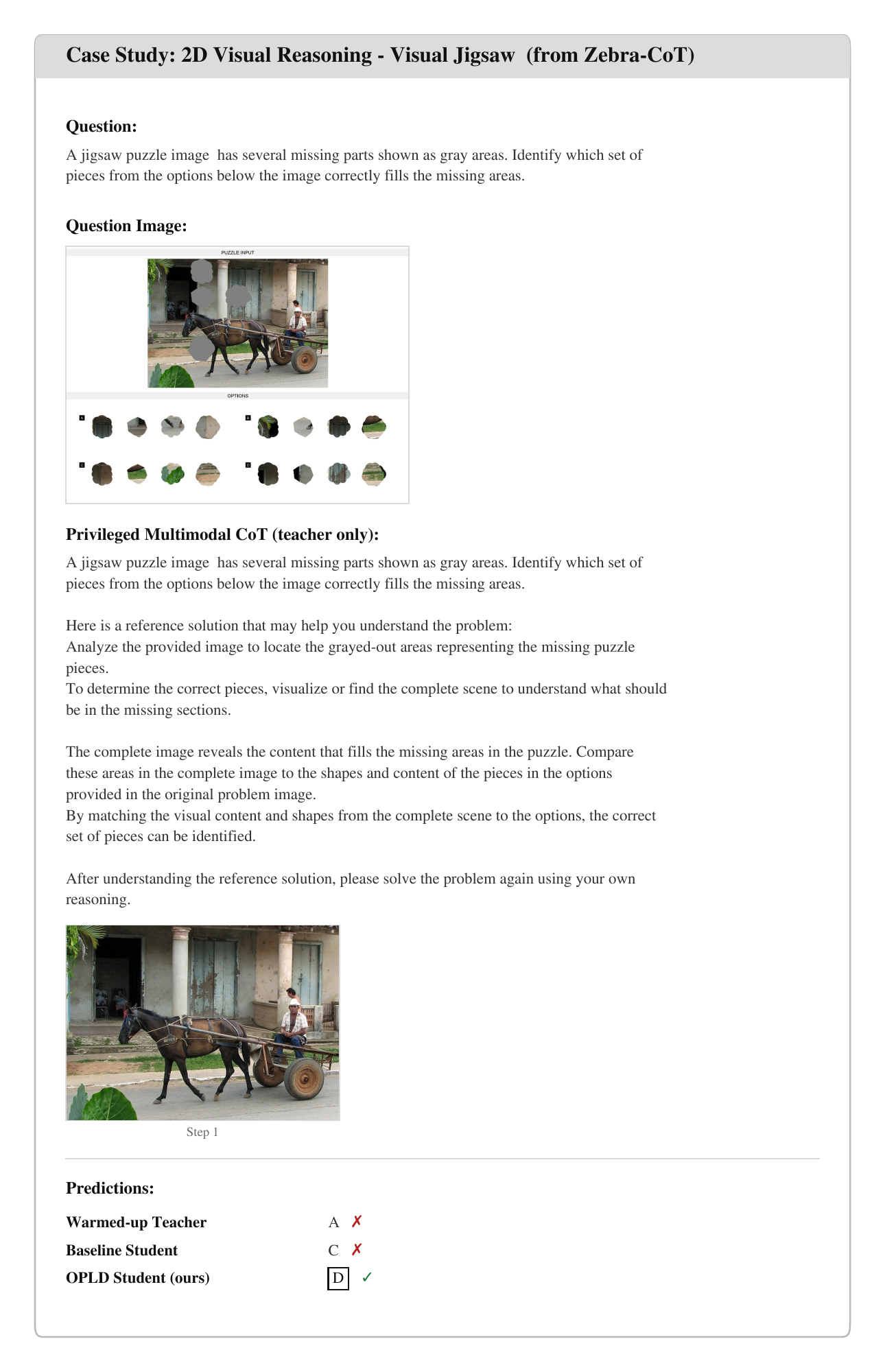}
\caption{
Inference Result
}
\end{figure*}

\begin{figure*}[t]
\centering
\includegraphics[width=0.7\textwidth]{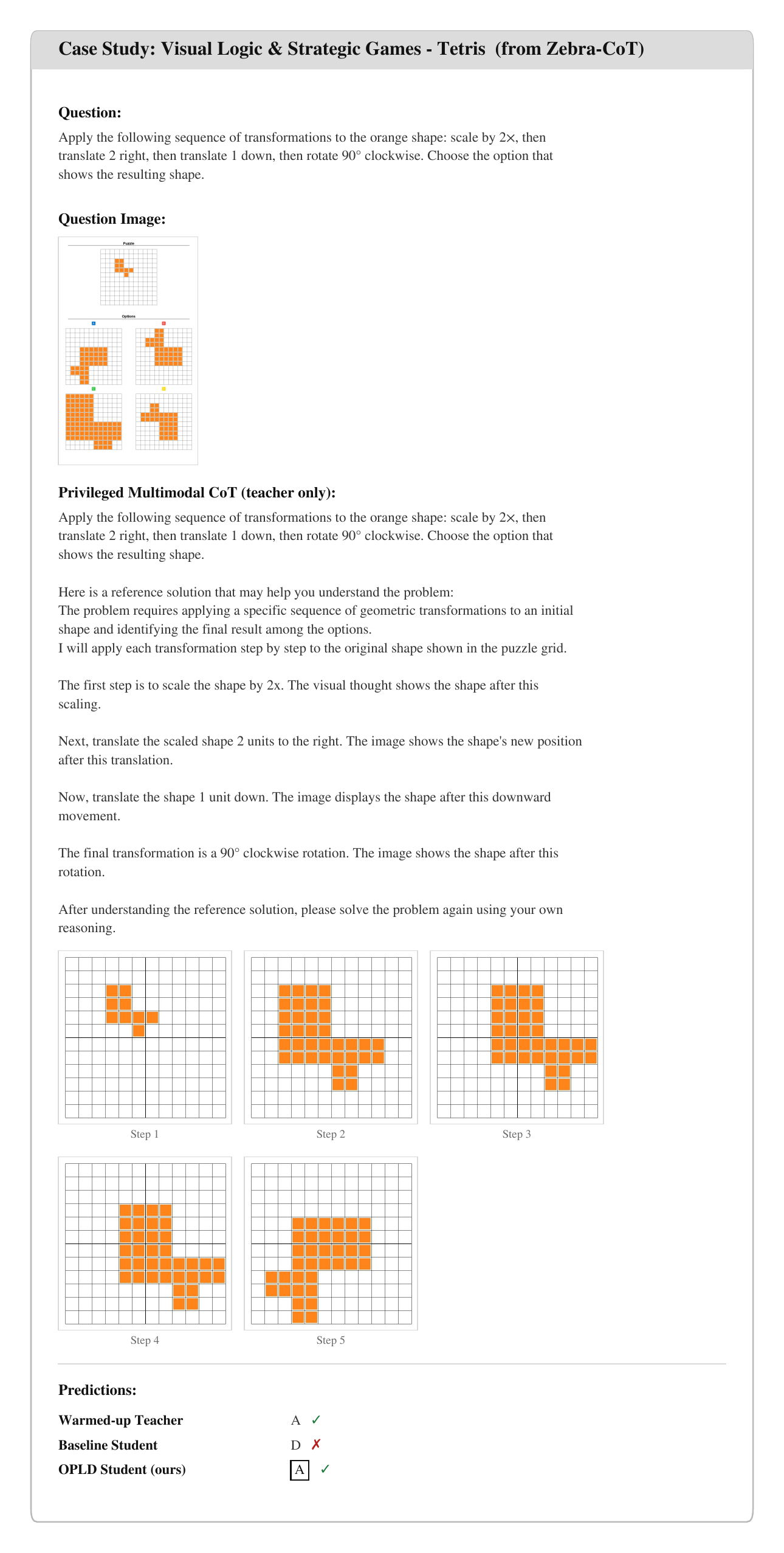}
\caption{
Inference Result
}
\end{figure*}


\begin{figure*}[t]
\centering
\includegraphics[width=0.9\textwidth]{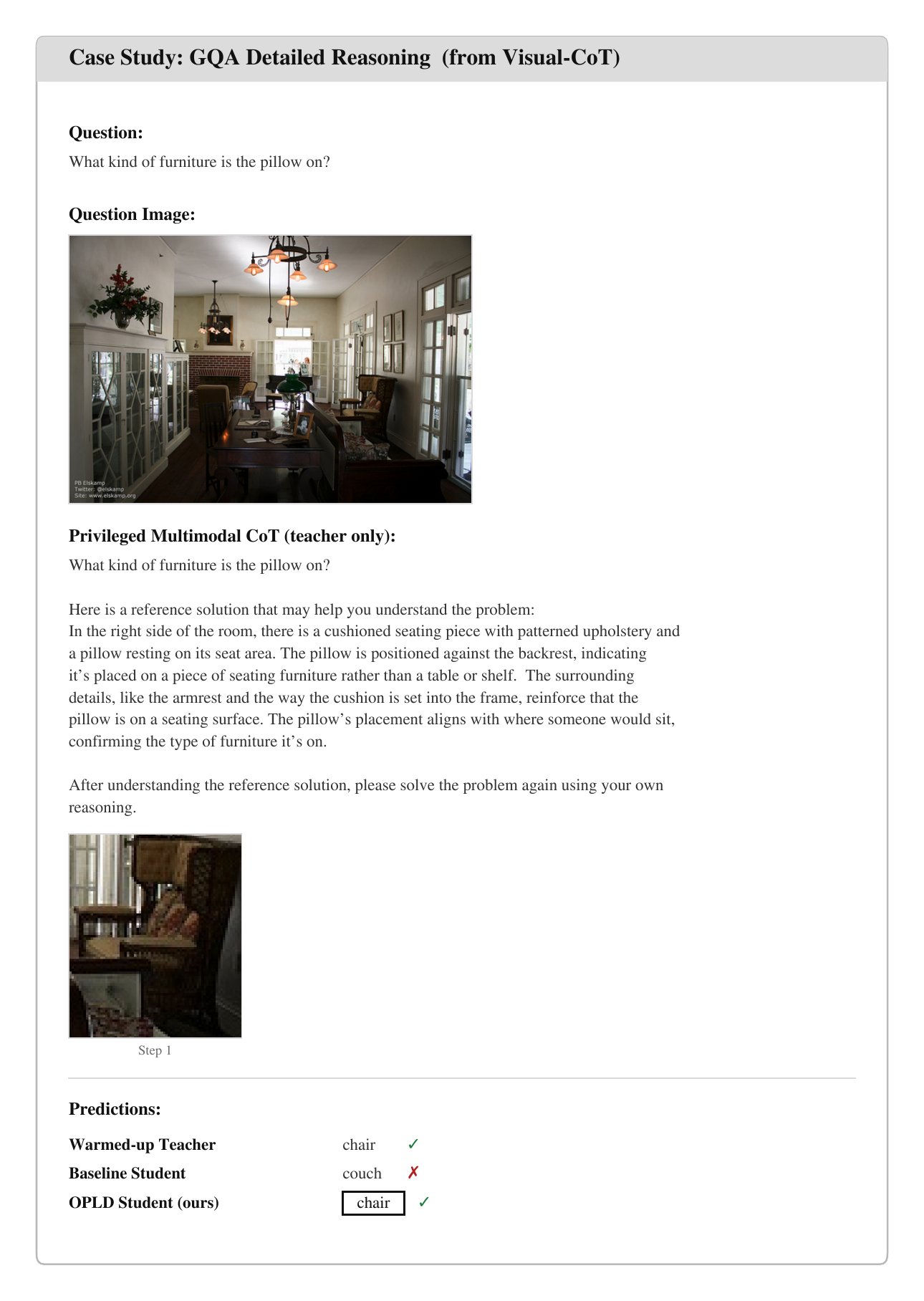}
\caption{
Inference Result
}
\end{figure*}

\begin{figure*}[t]
\centering
\includegraphics[width=0.9\textwidth]{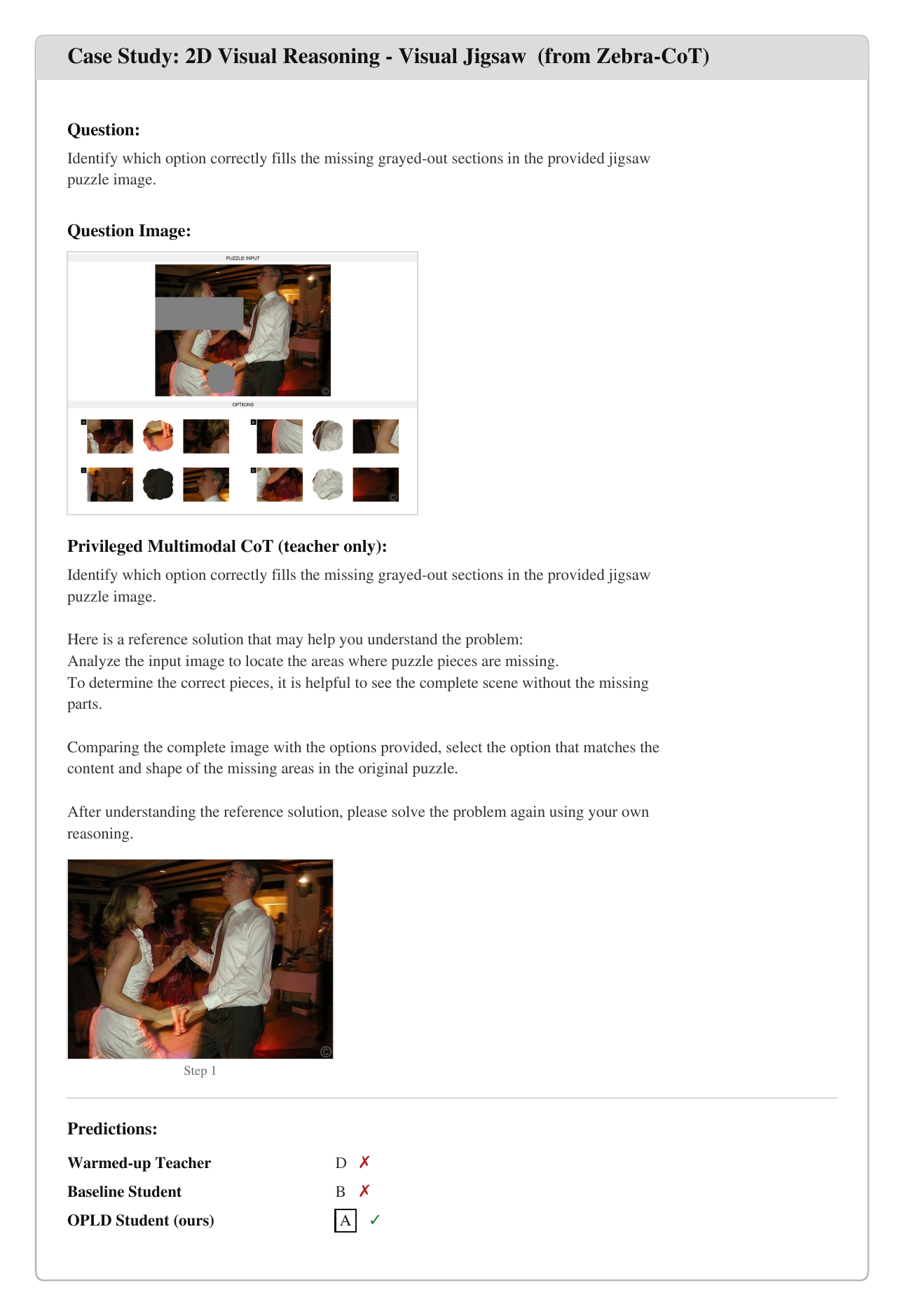}
\caption{
Inference Result
}
\end{figure*}

\end{document}